\documentclass[letterpaper]{article} 
\usepackage{aaai24}  
\usepackage{times}  
\usepackage{helvet}  
\usepackage{courier}  
\usepackage[hyphens]{url}  
\usepackage{graphicx} 
\usepackage{natbib}  
\usepackage{caption} 
\usepackage{algorithm}
\usepackage{algorithmic}
\usepackage{amsmath}
\usepackage{amssymb}
\usepackage{multirow}
\usepackage{subcaption}
\usepackage{makecell}
\usepackage{pifont}
\newcommand{\etal}{\textit{et al.}}
\newcommand{\cmark}{\ding{51}}
\newcommand{\xmark}{\ding{55}}

\usepackage{newfloat}
\usepackage{listings}
\DeclareCaptionStyle{ruled}{labelfont=normalfont,labelsep=colon,strut=off} 
\floatstyle{ruled}
\newfloat{listing}{tb}{lst}{}
\floatname{listing}{Listing}
\title{Self2Self+: Single-Image Denoising with Self-Supervised Learning and Image Quality Assessment Loss}
\author{
    Jaekyun Ko, Sanghwan Lee 
}
\affiliations{
    Department of Mechanical Convergence Engineering, Hanyang University, Korea\\
    \{rhworbs1124, shlee\}@hanyang.ac.kr
}

\begin{document}

\maketitle
\begin{abstract}
Recently, denoising methods based on supervised learning have exhibited promising performance. However, their reliance on external datasets containing noisy-clean image pairs restricts their applicability. To address this limitation, researchers have focused on training denoising networks using solely a set of noisy inputs. To improve the feasibility of denoising procedures, in this study, we proposed a single-image self-supervised learning method in which only the noisy input image is used for network training. Gated convolution was used for feature extraction and no-reference image quality assessment was used for guiding the training process. Moreover, the proposed method sampled instances from the input image dataset using Bernoulli sampling with a certain dropout rate for training. The corresponding result was produced by averaging the generated predictions from various instances of the trained network with dropouts. The experimental results indicated that the proposed method achieved state-of-the-art denoising performance on both synthetic and real-world datasets. This highlights the effectiveness and practicality of our method as a potential solution for various noise removal tasks.
\end{abstract}
\section{Introduction}
\label{sec:intro}
In image denoising, a clean image $x$ is recovered from a noisy image $y$. This operation is frequently modeled as $y = x + n$, where $n$ represents the measurement noise. Typically, $n$ is assumed to be additive white Gaussian noise (AWGN) with a standard deviation $\sigma$.

Neural networks (NNs) have demonstrated excellent performance in low-level vision tasks, including image denoising and super-resolution. The prevalent approach for training an image denoising NN involves using pairs of synthetic noisy observations $y_{i}$ and clean reference $x_{i}$. Using the trainable model parameter vector $\theta$ and a vast paired dataset, a denoising model $f_{\theta}$ was trained by solving the following equation:
\begin{equation} \label{eq:1}
  \theta^{*}={\operatorname*{argmin}_\theta}\sum_{i}\mathcal{L}({f_{\theta}(y_{i}),x_{i}}),
\end{equation}
where $\mathcal{L}(\cdot,\cdot)$ denotes the loss function that computes the distance between the two images. In this method, retaining abundant training samples yields a superior denoising performance; however, this process can occasionally be costly, challenging, or even unattainable.

To address this issue, self-supervised methods using only noisy external images have been introduced. In the Noise2Noise (N2N) method~\cite{n2n}, multiple pairs of two noisy images acquired from the same scene are used for training. The Noise2Void (N2V)~\cite{n2v} and Noise2Self (N2S)~\cite{n2s} methods adopt the \emph{blind-spot} strategy to train self-supervised models that prevent models from learning an identity mapping. However, this method is still not feasible for removing real-world noise satisfying the aforementioned prerequisites. First, acquiring numerous noisy images per incident or obtaining an external image that accurately represents the target noise distribution can be challenging and expensive. Second, regarding noise distribution, methods that rely on noise model assumptions often struggle to perform well. This is because the actual noise distribution in such images is typically unknown and can deviate significantly from the assumed noise model. Finally, the \emph{blind-spot} strategy incurs high computational cost, which poses a significant limitation for real-world applications.

Therefore, methods have been proposed that learn solely from the input image without any prerequisites to overcome these problems. Ulyanov \etal~\cite{dip} introduced the so-called \emph{deep image prior} (DIP), in which a deep learning model was used to extract an image prior. Xu \etal~\cite{nas} proposed a noise-as-clean strategy to resolve domain disparity in image priors and noise statistics between the training and test data. In the Self2Self (S2S) method~\cite{s2s}, Bernoulli sampling and dropout are used to improve performance and reduce prediction variance. 

In this study, we developed the Self2Self+ (S2S+), a self-supervised image-denoising framework inspired by Self2Self~\cite{s2s}, which is trained only on the input image. Specifically, using Bernoulli sampling and dropout~\cite{dropout} as the base scheme, we introduced a gated convolution (GConv)~\cite{gated_conv} and image quality assessment (IQA) loss function. Gated convolution learns a soft mask from the data and overcomes the partial convolution (PConv)~\cite{partial_conv} phenomenon, in which all channels in the same layer share an identical mask. Based on no-reference IQA (NR-IQA), the IQA loss function is adopted to maneuver the training of the deep learning model by minimizing the image quality score difference.

The contributions of our study are as follows:
\begin{itemize}
  \item A single-image self-supervised learning model was proposed to mitigate the limitation of requiring numerous training images consistent with the target data. 
  \item This method exhibited effective performance on both synthetic and real-world datasets, outperforming state-of-the-art single-image self-supervised methods. This indicates its strong denoising ability and generalization capability.
\end{itemize}
\begin{figure*}[h]
  \centering
  \includegraphics[width=\textwidth]{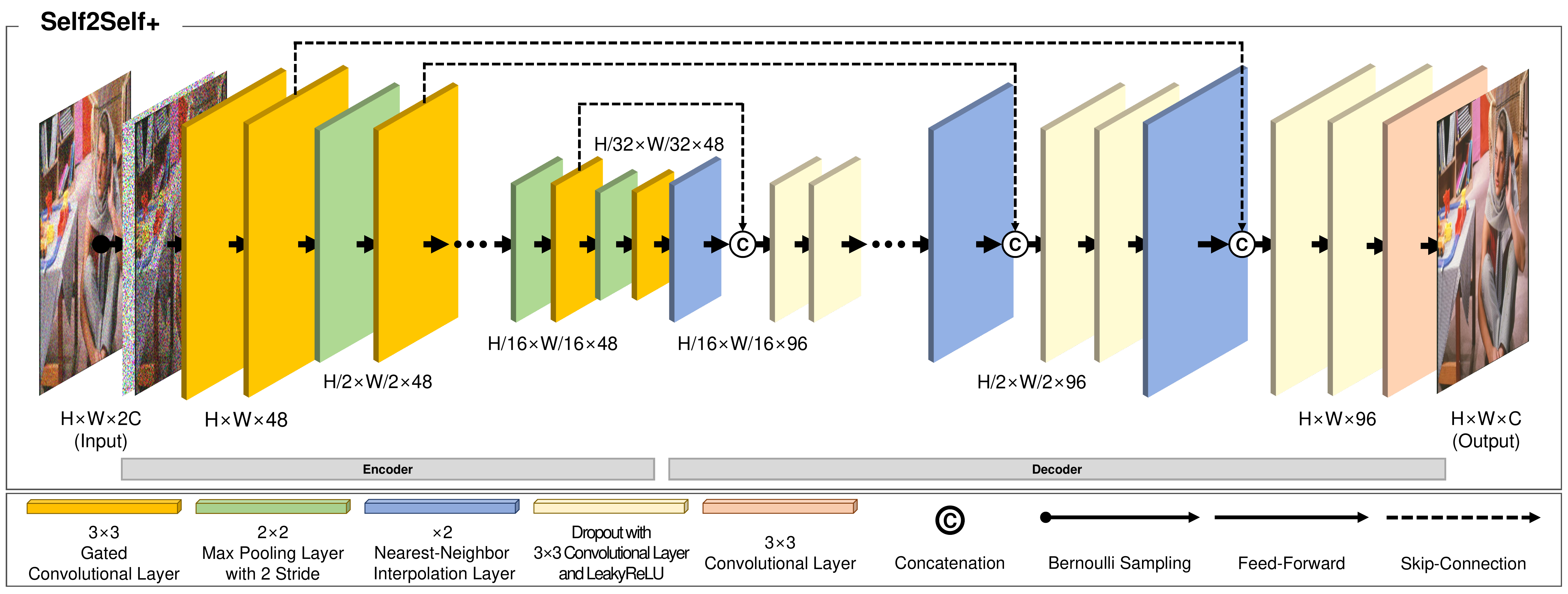}
  \caption{Proposed S2S+ architecture for single-image self-supervised learning.}
  \label{fig:model_architecture}
\end{figure*}
\section{Related Work}
\label{sec:related-work}
\subsection{Non-Learning Image Denoisers}
Various non-learning-based denoisers utilize predefined priors to model the noise distribution, which helps guide the denoising process. A commonly used prior in image denoising is the self-similarity prior obtained using non-local methods~\cite{nl_means, cbm3d, wmmn, mc_wmmn}. The \emph{NL-means}~\cite{nl_means} and CBM3D~\cite{cbm3d} methods exhibit excellent performance by presuming the existence of similar patches in a single image and using them in the noise removal process.
\subsection{Image Denoisers Trained with Noisy-Clean Image Pairs}
Image denoising methods based on NNs have rapidly evolved. Generally, a supervised learning method with a set of noisy-clean pairs is used for training. In the DnCNN method~\cite{dncnn}, a combination of an NN and residual learning strategy is used for image denoising, which considerably exceeds the denoising performance of non-learning methods. Since the introduction of the DnCNN method, a large number of NNs~\cite{rednet, memnet, cunet, n3_block, ffdnet, gcbd} have been developed to enhance the denoising accuracy.
\subsection{Image Denoisers Trained with Numerous Noisy Images}
Gathering noisy-clean pairs for supervised learning is difficult and expensive, which restricts the use of supervised denoisers. Therefore, the N2N method~\cite{n2n} uses pairs of two noisy images of the same incident for training, rather than using noisy-clean pairs. By assuming that the noise characteristics of the noisy-noisy pair are independent, the N2N method exhibits a similar performance to supervised denoising NNs. This is because the expectation of the mean squared error of the noisy-noisy pair is identical to that of the noisy-clean pair. However, assembling numerous noisy-noisy pairs remains challenging. Therefore, based on the \emph{blind-spot}, Krull \etal~\cite{n2v} proposed the N2V method, which only utilized unorganized noisy images for training. This method forces an NN to predict each pixel based on the adjacent pixels. Therefore, the method prevents the NN from learning an identity mapping, which is considered as a severe overfitting problem. Specifically, the loss is computed only on the image pixels that are replaced by the values of their randomly selected neighboring pixels. In the N2S method~\cite{n2s} and probabilistic extension of the N2V method (PN2V)~\cite{pn2v}, a framework similar to that of the N2V method was used. The external noisy images used for training should possess relevant content and noise statistics that are representative of the noise characteristic of the noisy image being restored.
\subsection{Image Denoisers Trained Only with A Single Noisy Image}
To address the abovementioned problem, denoising approaches with NNs using only noisy input images have been proposed. These networks represent the most adaptable approaches in real-world operations. In the DIP method~\cite{dip}, a generative NN is trained, in which a random input is projected onto a distorted image. This method assumes that priority over random sequences such as noise helps to understand useful image patterns. The S2S method~\cite{s2s} adopts Bernoulli-sampled instances of a noisy input image to produce multiple image pairs and avoid convergence to identity mapping. Moreover, dropout was used in both training and denoising stages to reduce the variance in the predictions. Inspired by the Neighbor2Neighbor method~\cite{neighbor2neighbor}, Lequyer \etal~\cite{n2f} introduced Noise2Fast (N2F) method, in which a novel down-sampling approach called \emph{chequerboard downsampling} was used to obtain a training set of four discrete images. This method trained the denoising NN by learning the mapping between adjacent pixels.
\subsection{Image Quality Assessment}
IQA plays a crucial role in computer vision tasks, serving as a reference indicator for evaluating image quality. IQA methods can be categorized into two main categories.

The first category is full-reference IQA (FR-IQA), in which a degraded image is compared with the original undistorted image for quality assessment. The strength of FR-IQA is that visual sensitivity can be computed using the difference between the degraded and reference images. This process adequately helps to mimic the behavior of the human visual system (HVS). The most common methods are the peak signal-to-noise ratio (PSNR) and structural similarity index (SSIM). To enhance the correlation with human perception, visual information fidelity (VIF)~\cite{vif}, multiscale SSIM (MS-SSIM)~\cite{ms_ssim}, feature similarity (FSIM)~\cite{fsim}, and learned perceptual image patch similarity (LPIPS)~\cite{lpips} have been introduced.

The second category is the no-reference IQA (NR-IQA) in which the quality of a degraded image is analyzed without using the original undistorted image. Compared with FR-IQA, NR-IQA mostly conducts the HVS process by relying on feature extraction. Thus, for this process, NNs are used to map the relationship between the image features and quality scores. Based on NNs, CNNIQA~\cite{cnniqa}, DeepIQA~\cite{deepiqa}, DBCNN~\cite{dbcnn}, and PaQ-2-PiQ~\cite{paq2piq} have been proposed.
\begin{table}[t]
\centering
\begin{tabular}{cccc}
\Xhline{2\arrayrulewidth}
\multicolumn{2}{c}{Dataset} & Probability & Training Step \\ \hline
\begin{tabular}[c]{@{}c@{}}Synthetic\\ noise\end{tabular}                   & CBSD68 & 0.4 & 4000 \\ \hline
\multirow{2}{*}{\begin{tabular}[c]{@{}c@{}}Real-world\\ noise\end{tabular}} & SIDD   & 0.9 & 1000 \\
           & PolyU          & 0.7         & 5000          \\ \Xhline{2\arrayrulewidth}
\end{tabular}%
\caption{Probability of the dropout layers and Bernoulli sampling, and number of training steps of our method selected for each dataset.}
\label{tab:hyperparameter}
\end{table}
\begin{figure*}[h]
\begin{subfigure}{.16\textwidth}
  \centering
  \includegraphics[width=.99\linewidth]{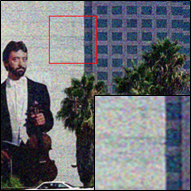}
  \caption{Input}
\end{subfigure}
\begin{subfigure}{.16\textwidth}
  \centering
  \includegraphics[width=.99\linewidth]{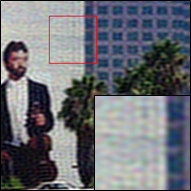}
  \caption{LPF}
\end{subfigure}
\begin{subfigure}{.16\textwidth}
  \centering
  \includegraphics[width=.99\linewidth]{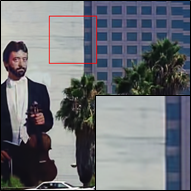}
  \caption{CBM3D}
\end{subfigure}
\begin{subfigure}{.16\textwidth}
  \centering
  \includegraphics[width=.99\linewidth]{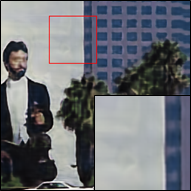}
  \caption{DIP}
\end{subfigure}
\begin{subfigure}{.16\textwidth}
  \centering
  \includegraphics[width=.99\linewidth]{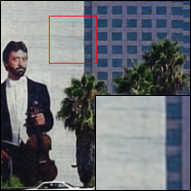}
  \caption{S2S}
\end{subfigure}
\begin{subfigure}{.16\textwidth}
  \centering
  \includegraphics[width=.99\linewidth]{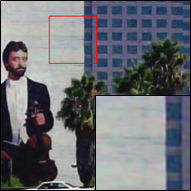}
  \caption{N2F}
\end{subfigure}\\
\begin{subfigure}{.16\textwidth}
  \centering
  \includegraphics[width=.99\linewidth]{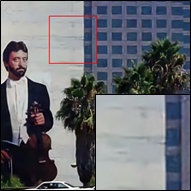}
  \caption{DNCNN}
\end{subfigure}
\begin{subfigure}{.16\textwidth}
  \centering
  \includegraphics[width=.99\linewidth]{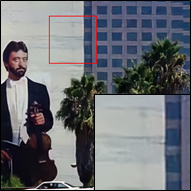}
  \caption{RED30}
\end{subfigure}
\begin{subfigure}{.16\textwidth}
  \centering
  \includegraphics[width=.99\linewidth]{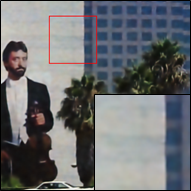}
  \caption{N2V}
\end{subfigure}
\begin{subfigure}{.16\textwidth}
  \centering
  \includegraphics[width=.99\linewidth]{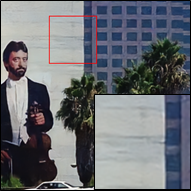}
  \caption{N2N}
\end{subfigure}
\begin{subfigure}{.16\textwidth}
  \centering
  \includegraphics[width=.99\linewidth]{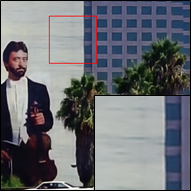}
  \caption{Ours}
\end{subfigure}
\begin{subfigure}{.16\textwidth}
  \centering
  \includegraphics[width=.99\linewidth]{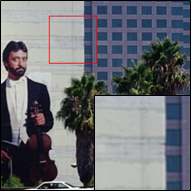}
  \caption{GT}
\end{subfigure}
\caption{Qualitative results of our method and other baselines on \emph{CBSD68} corrupted by AWGN with a noise level $\sigma=25$.}
\label{fig:cbsd68_evaluation}
\end{figure*}
\begin{table*}[t]
\centering
\begin{tabular}{c|cccccccccc}
\Xhline{2\arrayrulewidth}
\hline
\multirow{2}{*}{Methods} & \multicolumn{6}{c|}{Non-learning or single-image learning methods} & \multicolumn{4}{c}{Dataset-based deep learning methods} \\ \cline{2-11}
            & LPF    & CBM3D  & DIP    & S2S    & N2F    & \multicolumn{1}{c|}{Ours}            & DnCNN  & RED30  & N2V    & N2N    \\ \hline
Noise Level & \multicolumn{10}{c}{PSNR (dB)}                                                                                        \\ \hline
$\sigma=15$ & 24.01  & 33.20  & 27.06  & 28.18  & 28.00  & \multicolumn{1}{c|}{\textbf{29.97}}  & 33.10  & 33.38  & 27.34  & 33.27  \\
$\sigma=25$ & 23.38  & 30.16  & 26.69  & 27.24  & 26.68  & \multicolumn{1}{c|}{\textbf{28.71}}  & 30.68  & 30.87  & 26.61  & 30.42  \\
$\sigma=50$ & 21.46  & 25.86  & 24.76  & 24.39  & 24.07  & \multicolumn{1}{c|}{\textbf{26.51}}  & 27.62  & 27.80  & 24.82  & 26.49  \\ \hline
Noise Level & \multicolumn{10}{c}{SSIM}                                                                                             \\ \hline
$\sigma=15$ & 0.7846 & 0.9592 & 0.8638 & 0.9125 & 0.9008 & \multicolumn{1}{c|}{\textbf{0.9307}} & 0.9595 & 0.9614 & 0.884  & 0.962  \\
$\sigma=25$ & 0.7421 & 0.9251 & 0.8552 & 0.8857 & 0.8631 & \multicolumn{1}{c|}{\textbf{0.9076}} & 0.9325 & 0.9356 & 0.8514 & 0.9328 \\
$\sigma=50$ & 0.6205 & 0.8399 & 0.8064 & 0.786  & 0.7773 & \multicolumn{1}{c|}{\textbf{0.8557}} & 0.8774 & 0.8831 & 0.7916 & 0.8634 \\ \hline \Xhline{2\arrayrulewidth}
\end{tabular}%
\caption{Quantitative results of our method and other baselines on \emph{CBSD68} corrupted by AWGN with noise levels $\sigma=\{15, 25,50\}$. Our results are marked in \textbf{bold}.}
\label{tab:cbsd68_evaluation}
\end{table*}
\section{Methodology}
\label{sec:methodology}
In this section, we first introduce the architecture of the proposed method and then present a detailed explanation of the single-image self-supervised training and denoising schemes.
\subsection{Neural Network Architecture}
The NN structure of the proposed method, wherein an autoencoder framework is employed, is shown in Figure \ref{fig:model_architecture}. Given a noisy input image of size ${H}\times{W}\times{C}$, the generated mask is concatenated with the input image to obtain the spatial information of the missing pixels, yielding an input of size ${H}\times{W}\times{2C}$. The encoder first expands the number of input channels to ${H}\times{W}\times{48}$ using two GConv~\cite{gated_conv} layers, wherein LeakyReLU (LReLU) is used as the activation function. The feature map is then down-sampled using a max-pooling layer and input into a subsequent GConv layer. This process continues until the final output of the encoder reaches $\frac{H}{32}\times\frac{W}{32}\times{48}$. The number of channels for all outputs in the encoder is fixed at 48.

A combination of nearest-neighbor interpolation with a scaling factor of two, a concatenation operation, and two vanilla convolutional layers with LReLUs are continuously used in the decoder. The concatenation operation accepts the up-sampled feature map and the output of the encoder, which has the same spatial size as the output of the decoder. Except for the last layer, all convolutional layers in the decoder are combined with dropout and have 96 output channels. For the final output, a single vanilla convolutional layer without dropout and LReLU transforms the feature map into ${H}\times{W}\times{C}$ to match the input image size. 
\subsection{Training Scheme}
To prevent the NN from converging on an identity mapping, a group of numerous Bernoulli-sampled instances of a noisy image $y$ was generated. With instances of $y$ defined as $\{\hat{y}_{m}\}^{M}_{m=1}$, the sampled instance is computed as follows:
\begin{equation}
    \hat{y}:={b}\odot{y},
\end{equation} \label{eq:2}
where $b$ denotes a binary Bernoulli vector instance, in which a Bernoulli distribution with $p\in(0,1)$ is used to independently sample the pixel values, and $\odot$ represents element-wise multiplication. Each $m$ in a group of image pairs $\{(\hat{y}_{m},\Bar{y}_{m})\}^{M}_{m=1}$ is formulated as follows:
\begin{equation} \label{eq:3}
    \hat{y}_{m}:={b}_{m}\odot{y}; \qquad \Bar{y}_{m}:=(1-{b}_{m})\odot{y}.
\end{equation}
Here, a set of image pairs was used to train the NN by minimizing the corresponding self-supervised loss function.
\begin{equation} \label{eq:4}
  \mathcal{L}_{self-supervised}={\sum^{M}_{m=1}}{\|f_{\theta}(\hat{y}_{m})-\Bar{y}_{m}\|^{1}_{b_{m}}},
\end{equation}
where $\theta$ denotes the network parameter vector and $f_{\theta}$ is the denoising NN. Unlike the S2S method~\cite{s2s}, we used the sum absolute error (SAE) instead of the sum squared error (SSE) to compute the loss, where $\|\cdot\|^{1}_{b}$ is equal to $\|(1-b)\odot\cdot\|^{1}_{1}$. The aggregation of the loss function over the entire pair calculates the difference over all image pixel values owing to the random selection of masked pixels using a Bernoulli process.

Moreover, assuming that the noise components are independent and have zero mean, the expectation of the loss function defined in Eq. (\ref{eq:4}) with respect to noise $n$ for an arbitrary $f_{\theta}$ is expressed as follows:
\begin{equation} \label{eq:5}
    \mathbb{E}_{n}\left[{\sum^{M}_{m=1}}{\|f_{\theta}(\hat{y}_{m})-\Bar{y}_{m}\|^{1}_{b_{m}}}\right]=\begin{cases} 
    {\sum^{M}_{m=1}}{\|f_{\theta}(\hat{y}_{m})-x\|^{1}_{b_{m}}}, \\ \text{$if {\;} f_{\theta}(\hat{y}_{m})>y$}\\ \\
    {\sum^{M}_{m=1}}{\|x-f_{\theta}(\hat{y}_{m})\|^{1}_{b_{m}}}. \\ \text{$otherwise$}
    \end{cases}
\end{equation}
Using the pairs of Bernoulli-sampled instances $\{\hat{y}_{m},\Bar{y}_{m}\}$ is similar to using the pairs of sampled instances $\hat{y}_{m}$ and cleaning reference $x$ to train the NN with sufficient pairs. Please refer to the supplementary material for further details.

In addition to the self-supervised loss function, we introduced an IQA loss function to improve the image restoration process. Because NR-IQA methods only use input images to compute quality scores, they can be used as IQA loss functions for single-image learning tasks. Specifically, we used the PaQ-2-PiQ~\cite{paq2piq} method, pretrained with the KonIQ-10k dataset~\cite{koniq10k}. To use the local and global features of the input image, this method cropped multiple patches of various sizes from the image to calculate the mean opinion score (MOS), which was normalized to a range of 0 to 100. Because images with satisfactory qualities exhibit a higher MOS, the IQA loss function computes the SSE between the predicted score of a denoised image and 100. Thus, using the NN implemented in the PaQ-2-PiQ method, denoted by $f_{P2P}$, the IQA loss function is expressed as the following equation.
\begin{equation}
    \mathcal{L}_{IQA}={\sum^{M}_{m=1}}\|100-f_{P2P}(f_{\theta}(\hat{y}_{m}))\|^{2}_{2}.
\end{equation}

Our full loss function is a combination of $\mathcal{L}_{self-supervised}$ and $\mathcal{L}_{IQA}$, which is formulated as follows:
\begin{equation}
    \mathcal{L}=\mathcal{L}_{self-supervised}+\lambda_{IQA}\mathcal{L}_{IQA},
\end{equation}
where the weight of the IQA loss function was empirically chosen as given below. 
$\lambda_{IQA}={2}\times{10^{-8}}$.
\begin{figure*}[h]
\centering
\begin{subfigure}{.245\textwidth}
  \centering
  \includegraphics[width=.99\linewidth]{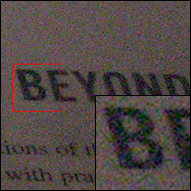}
  \caption{Input}
\end{subfigure}
\begin{subfigure}{.245\textwidth}
  \centering
  \includegraphics[width=.99\linewidth]{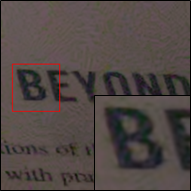}
  \caption{CBM3D}
\end{subfigure}
\begin{subfigure}{.245\textwidth}
  \centering
  \includegraphics[width=.99\linewidth]{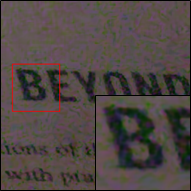}
  \caption{DIP}
\end{subfigure}
\begin{subfigure}{.245\textwidth}
  \centering
  \includegraphics[width=.99\linewidth]{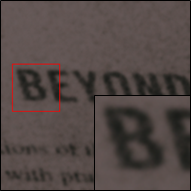}
  \caption{S2S}
\end{subfigure}\\
\begin{subfigure}{.245\textwidth}
  \centering
  \includegraphics[width=.99\linewidth]{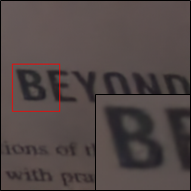}
  \caption{RED30}
\end{subfigure}
\begin{subfigure}{.245\textwidth}
  \centering
  \includegraphics[width=.99\linewidth]{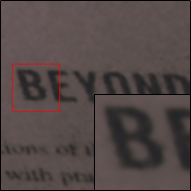}
  \caption{Ours}
\end{subfigure}
\begin{subfigure}{.245\textwidth}
  \centering
  \includegraphics[width=.99\linewidth]{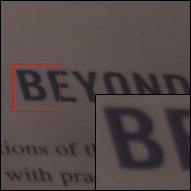}
  \caption{GT}
\end{subfigure}
\caption{Qualitative results of our method and other baselines on \emph{SIDD}.}
\label{fig:sidd_evaluation}
\end{figure*}

\begin{figure*}[h]
\centering
\begin{subfigure}{.245\textwidth}
  \centering
  \includegraphics[width=.99\linewidth]{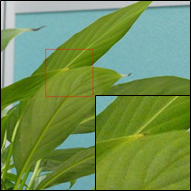}
  \caption{Input}
\end{subfigure}
\begin{subfigure}{.245\textwidth}
  \centering
  \includegraphics[width=.99\linewidth]{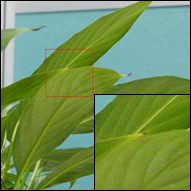}
  \caption{CBM3D}
\end{subfigure}
\begin{subfigure}{.245\textwidth}
  \centering
  \includegraphics[width=.99\linewidth]{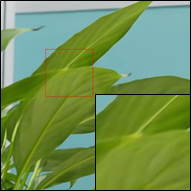}
  \caption{DIP}
\end{subfigure}
\begin{subfigure}{.245\textwidth}
  \centering
  \includegraphics[width=.99\linewidth]{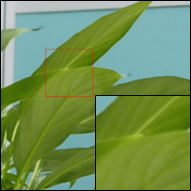}
  \caption{S2S}
\end{subfigure}\\
\begin{subfigure}{.245\textwidth}
  \centering
  \includegraphics[width=.99\linewidth]{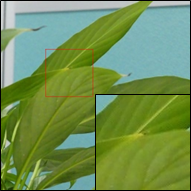}
  \caption{RED30}
\end{subfigure}
\begin{subfigure}{.245\textwidth}
  \centering
  \includegraphics[width=.99\linewidth]{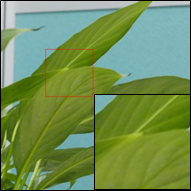}
  \caption{Ours}
\end{subfigure}
\begin{subfigure}{.245\textwidth}
  \centering
  \includegraphics[width=.99\linewidth]{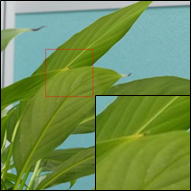}
  \caption{GT}
\end{subfigure}
\caption{Qualitative results of our method and other baselines on \emph{PolyU}.}
\label{fig:polyu_evaluation} 
\end{figure*}
\subsection{Denoising Scheme}
Similar to the S2S method, we adopted dropout layers in the denoising NN to reduce prediction variance. Because dropout is still valid at the denoising stage, various NNs can be produced from the trained NN, which provides estimators with a certain measure of independence.

In the denoising stage, initiating a dropout on the constructed layers of the trained NN $f_{\theta_{*}}$ creates different NNs, denoted by $f_{\theta_{1}},\cdots,f_{\theta_{N}}$. Furthermore, the Bernoulli-sampled instances of $y$ are provided as the input of the generated NNs to obtain the restored images $\Hat{x}_{1},\cdots,\Hat{x}_{N}$. To acquire the final reconstructed image $x^{*}$, the restored images were averaged by the following computation:
\begin{equation}
    x^{*}=\frac{1}{N}\sum^{N}_{n=1}\Hat{x}_{n}=\frac{1}{N}\sum^{N}_{n=1}f_{\theta_{n}}(b_{M+n}\odot{y}).
\end{equation}
In this method, we used 500 Bernoulli-sampled instances to obtain the final recovered image.
\section{Experiment}
\label{sec:experiment}
In this section, the implementation details are presented. Then, the proposed method is extensively evaluated by utilizing state-of-the-art denoising methods with synthetic and real-world noisy images. Finally, an ablation study is conducted to evaluate the effectiveness of GConv and introduced loss functions.
\subsection{Implementation Details}
Throughout the experiments, all GConv and vanilla convolutional layers had a kernel size of ${3}\times{3}$, strides of 1, and reflection padding of length 2 to reduce artifacts. The hyperparameter of each LReLU was set to 0.2. We used the Adam~\cite{adam} optimizer with a learning rate initialized to ${4}\times{10^{-4}}$. The probabilities of dropout layers and Bernoulli sampling, as well as the training steps selected for each dataset, are listed in Table \ref{tab:hyperparameter}. The proposed method was implemented using the PyTorch~\cite{pytorch} framework and an NVIDIA GEFORCE RTX 3090 Ti GPU.
\subsection{Evaluation on Synthetic Noise Removal} \label{eval_synthetic_noise}
For synthetic noise experiments, we used the DIV2K dataset~\cite{div2k}, which contained 800 images with 2K resolution, to train dataset-based deep learning methods. To develop a training dataset, we randomly cropped ${256}\times{256}$ patches and added an AWGN with noise levels $\sigma=\{15,25,50\}$ to all patches. For testing, we used the CBSD68 dataset used in~\cite{cbsd68} with 68 images. For the baseline comparisons, we selected non-learning or single-image learning methods, that are, low-pass filtering (LPF), CMB3D~\cite{cbm3d}, DIP~\cite{dip}, S2S~\cite{s2s}, and N2F~\cite{n2f}, and dataset-based methods, namely, DnCNN~\cite{dncnn}, RED30~\cite{rednet}, N2V~\cite{n2v}, and N2N~\cite{n2n}, for performance comparison. As illustrated in Figure \ref{fig:cbsd68_evaluation}, non-learning and single-image learning methods leave residual noise on the wall and window chassis. Notably, the DIP method is unable to retain texture because of excessive blurring. However, the proposed method effectively removed noise while preserving structures, particularly compared with the S2S method. As listed in Table \ref{tab:cbsd68_evaluation}, our method surpassed the performance of non-learning or single-image learning methods, including, LPF, DIP, S2S, and N2F, and N2V, a dataset-based deep learning method, by at least 2.24 dB in PSNR and 0.0406 in SSIM. Furthermore, when considering a noise level of $\sigma = 50$, our method outperformed the CBM3D and N2N methods, which demonstrates the denoising robustness of the proposed method against strong noise levels. We speculated that using GConv, SAE in the self-supervised loss function induces effective noise removal, whereas IQA loss function helps to retain the overall texture. Please refer to the supplementary material for more qualitative results.
\begin{table*}[t]
\centering
\begin{tabular}{c|ccccc}
\Xhline{2\arrayrulewidth}
\hline
 &
  \multicolumn{5}{c}{SIDD Dataset} \\ \hline
\multirow{2}{*}{Methods} &
  \multicolumn{4}{c|}{Non-learning or single-image learning methods} &
  Dataset-based deep learning methods \\ \cline{2-6} 
 &
  \multicolumn{1}{c|}{CBM3D} &
  \multicolumn{1}{c|}{DIP} &
  \multicolumn{1}{c|}{S2S} &
  \multicolumn{1}{c|}{Ours} &
  RED30 \\ \hline
PSNR (dB) &
  \multicolumn{1}{c|}{31.71} &
  \multicolumn{1}{c|}{29.43} &
  \multicolumn{1}{c|}{30.84} &
  \multicolumn{1}{c|}{\textbf{34.11}} &
  38.61 \\ \hline
SSIM &
  \multicolumn{1}{c|}{0.7825} &
  \multicolumn{1}{c|}{0.7338} &
  \multicolumn{1}{c|}{0.7263} &
  \multicolumn{1}{c|}{\textbf{0.8903}} &
  0.9517 \\ \hline
 &
  \multicolumn{5}{c}{PolyU Dataset} \\ \hline
\multirow{2}{*}{Methods} &
  \multicolumn{4}{c|}{Non-learning or single-image learning methods} &
  Dataset-based deep learning methods \\ \cline{2-6} 
 &
  \multicolumn{1}{c|}{CBM3D} &
  \multicolumn{1}{c|}{DIP} &
  \multicolumn{1}{c|}{S2S} &
  \multicolumn{1}{c|}{Ours} &
  RED30 \\ \hline
PSNR (dB) &
  \multicolumn{1}{c|}{36.74} &
  \multicolumn{1}{c|}{36.06} &
  \multicolumn{1}{c|}{35.30} &
  \multicolumn{1}{c|}{\textbf{37.33}} &
  37.65 \\ \hline
SSIM &
  \multicolumn{1}{c|}{0.9755} &
  \multicolumn{1}{c|}{0.9785} &
  \multicolumn{1}{c|}{0.9472} &
  \multicolumn{1}{c|}{\textbf{0.9815}} &
  0.9823 \\ \hline \Xhline{2\arrayrulewidth}
\end{tabular}%
\caption{Quantitative results of our method and other baselines on \emph{SIDD} and \emph{PolyU}. Our results are marked in \textbf{bold}.}
\label{tab:real_world_evaluation}
\end{table*}
\subsection{Evaluation on Real-World Noise Removal} \label{eval_real_noise}
For the real-world noise experiments, we used two real-world datasets with CMB3D~\cite{cbm3d}, DIP~\cite{dip}, S2S~\cite{s2s}, and RED30~\cite{rednet} selected as the baseline methods for performance comparisons.

First, we used the SIDD dataset~\cite{sidd}, which was compiled using five smartphone cameras for ten static scenes. For the dataset-based deep learning methods, we used the SIDD-Medium dataset, which encompasses 320 pairs of noisy and corresponding reference images with 4K or 5K resolutions. All images were randomly cropped to generate ${256}\times{256}$ patches. For testing, 1280 cropped patches of size ${256}\times{256}$ were collected from the SIDD validation dataset. As shown in Figure \ref{fig:sidd_evaluation}, all baselines, except for RED30, exhibited residual noise and artifacts. In particular, S2S did not sufficiently eliminate noise, and the resulting blur degraded the quality of letters in the output. In contrast, the proposed method eliminated intense noise while accurately maintaining the colors and structures of the letters. Please refer to the supplementary material for more qualitative results.

Next, we utilized the PolyU dataset~\cite{polyu}, which comprises 40 different static scenes captured by five cameras. We cropped 100 regions of ${512}\times{512}$ from these scenes and randomly selected 70 images cropped to ${256}\times{256}$ patches to train the dataset-based deep learning methods. The remaining 30 images were used for testing. As shown in Figure \ref{fig:polyu_evaluation}, all baselines, apart from RED30, left residual noise or excessively smoothed the images, with the S2S method producing the least distinct lines on the leaves. In contrast, our method depicted sharp lines with a well-preserved leaf texture. Please refer to the supplementary material for more qualitative results.

The quantitative metrics of both datasets are reported in Table \ref{tab:real_world_evaluation}, which compares our method’s generalized denoising performance with that of several baselines.
\subsection{Ablation Study}
\begin{table}[t]
\centering
\resizebox{\columnwidth}{!}{%
\begin{tabular}{cccc|cc}
\hline \Xhline{2\arrayrulewidth}
GConv & $L_{IQA}$ & $L_{1-ss}$ & $L_{2-ss}$ & PSNR (dB) & SSIM   \\ \hline
\xmark & \xmark & \cmark & \xmark & 26.70     & 0.8534 \\
\cmark & \xmark & \cmark & \cmark & 26.96     & 0.8642 \\
\cmark & \cmark & \cmark & \xmark & \textbf{27.00} & \textbf{0.8653} \\
\cmark & \cmark & \xmark & \cmark & 25.75     & 0.8401 \\ \hline \Xhline{2\arrayrulewidth}
\end{tabular}%
}
\caption{Ablation study on \emph{CBSD68} corrupted by AWGN with a noise level $\sigma=50$. GConv: gated convolution. $L_{IQA}$: IQA loss function. $L_{1-ss}$: SAE in the self-supervised loss function. $L_{2-ss}$: SSE in the self-supervised loss function. The best results are marked in \textbf{bold}.}
\label{tab:ablation_study}
\end{table}
To determine the effectiveness of the novel methodologies implemented with S2S+, we performed an ablation study for a comprehensive analysis. In particular, we considered the effect of GConv, IQA loss function, and dimensions of the self-supervised loss function. First, to analyze the effectiveness of GConv, we integrated only GConv into the base network. As presented in Table \ref{tab:ablation_study}, PSNR and SSIM are enhanced by 0.23 dB and 0.0108, respectively, as the vanilla convolution is replaced by the GConv. This phenomenon indicates that filling in missing pixel values by adopting GConv can be used to predict appropriate values and reduce artifacts using the prior spatial information from the generated mask. Subsequently, we added $L_{IQA}$ to the self-supervised loss function to verify its effectiveness. The results with $L_{IQA}$ achieved superior outcomes for both PSNR and SSIM by 0.04 dB and 0.0011, respectively. Therefore, integrating $L_{IQA}$ in training reduces the MOS difference distance. Moreover, this phenomenon encourages the NN to obtain more acceptable results for HVS, which eventually increases the values of FR-IQA metrics. Finally, the dimensions of $\mathcal{L}_{self-supervised}$ are examined. When SSE, $L_{2-ss}$, is used instead of SAE, $L_{1-ss}$, PSNR and SSIM both decrease considerably by 1.25 dB and 0.0252, respectively, which corresponded to the lowest outcome. Thus, we speculate that using $L_{1-ss}$ in our method prevents excessive blurring and improves high-frequency restoration.
\section{Conclusion}
\label{sec:conclusion}
We proposed a novel single-image self-supervised deep learning method. The method does not require any prerequisites to construct the training dataset because only a noisy input image is used for training. Based on the S2S method~\cite{s2s}, we used a dropout-based scheme in both the training and denoising stages to reduce the variance of the prediction and prevent overfitting. Moreover, we used gated convolution at the encoder to replace the removed pixels with appropriate values by learning a soft mask. We also introduced the IQA loss function to generate results that were more adequate for HVS. Experiments on synthetic and real-world noise removal indicate that the proposed method outperforms other non-learning or single-image learning-based methods and produced images with less residual noise and fewer artifacts. Therefore, the proposed method is a promising solution for practical denoising problems.

\pagebreak
\section{Additions to section 3: Methodology}
\subsection{Gated Convolution}
Generally, vanilla convolutions are used for image inpainting tasks ~\cite{glcic, contextual_attention} to fill in missing regions such that the generated results are visually and semantically acceptable. Given the input feature map with the \emph{C-channel}, each pixel at $(i,j)$ in the output map with the \emph{C'-channel} of the vanilla convolution is formulated as follows:
\begin{equation}
    O_{i,j}=\sum_{{\Delta}i=-k'_{h}}^{k'_{h}}\sum_{{\Delta}j=-k'_{w}}^{k'_{w}}{W}_{k'_{h}+{\Delta}i,k'_{w}+{\Delta}j}\; {\cdot}\; {I}_{i+{\Delta}i,j+{\Delta}j},
\end{equation}
where $j$, $i$ denote x-axis, y-axis of the output map, $k_{h}$, $k_{w}$ represent the kernel size, $k'_{h}=\frac{k_{h}-1}{2}$, $k'_{w}=\frac{k_{w}-1}{2}$, $W$ are convolutional filters and ${I}_{i+{\Delta}i,j+{\Delta}j}$, $O_{i,j}$ represent the input and output feature map. However, visual artifacts, such as blurriness, color inconsistency, and unnatural responses besieging holes, are induced because vanilla convolutions use the same filters on all valid, invalid, and mixed pixels. 

To mitigate this problem, partial convolution (PConv)~\cite{partial_conv} was introduced, in which a masking and re-normalization process were adopted to force the convolution to be dependent only on valid pixels. Therefore, PConv was computed as follows: 
\begin{equation}
    O_{i,j}=\begin{cases}
   \sum\sum{W}\cdot({I}\odot\frac{M}{sum(M)}), \quad & \text{$if {\;} sum(M)>0$}\\
    0, \quad & \text{$otherwise$}
    \end{cases}
\end{equation}
where $M$ is a binary mask in which 0 and 1 denote the invalid and valid pixels at location $(i,j)$, respectively, and $\odot$ represents element-wise multiplication. The following rule is applied to $M$ for refurbishment after each PConv operation: $m'_{i,j}=1$ and $iif {\;} sum(M)>0$. However, PConv still has some limitations of invalid pixels in $M$ are steadily transformed into a value of 1 as the layers deepen, and all channels in the same layer share an identical mask.

Yu \etal~\cite{gated_conv} introduced gated convolution (GConv) to learn soft masks from data instead of using unlearnable hard-gating mask, as in PConv. GConv is expressed as follows:
\begin{equation}
    \begin{aligned}
    &Gating_{i,j}=\sum\sum{W_{g}}\cdot{I},\\
    &Feature_{i,j}=\sum\sum{W_{f}}\cdot{I},\\
    &O_{i,j}=\phi(Feature_{i,j})\odot\sigma(Gating_{i,j}),
    \end{aligned}
\end{equation}
where $W_{g}$ and $W_{f}$ represent two convolutional filters, $\sigma$ is a sigmoid function used to project output values between zero and one, and $\phi$ denotes an activation function.

Because the pixels of the noisy input image are randomly masked using Bernoulli sampling, the NN must fill in missing values, as in the image inpainting task, to restore the image. Therefore, GConv is suitable for our method, and we used it only at the encoder, because the NN can fill in most of pixel values in the encoding process. Moreover, using vanilla convolutions in the decoder reduces the number of model parameters and prevents the overfitting problem. 
\subsection{Proof on Self-Supervised Loss Function}
The self-supervised loss function can be rewritten as follows:
\begin{equation} \label{eq:12}
    \sum^{M}_{m=1}{\|f_{\theta}(\hat{y}_{m})-\Bar{y}_{m}\|^{1}_{b_{m}}}=\sum^{M}_{m=1}{\|f_{\theta}(\hat{y}_{m})-y\|^{1}_{b_{m}}}.
\end{equation}
If $f_{\theta}(\hat{y}_{m})>y$, Eq. (\ref{eq:12}) can be expressed as follows:
\begin{equation} \label{eq:13}
    \begin{aligned}
    &\sum^{M}_{m=1}{\|f_{\theta}(\hat{y}_{m})-y\|^{1}_{b_{m}}}=\\
    &\sum^{M}_{m=1}{\|f_{\theta}(\hat{y}_{m})-(x+n)\|^{1}_{b_{m}}}=\\
    &\sum^{M}_{m=1}{\|f_{\theta}(\hat{y}_{m})-x\|^{1}_{b_{m}}}-\sum^{M}_{m=1}{\|n\|^{1}_{b_{m}}}.
    \end{aligned}
\end{equation}
However, if $f_{\theta}(\hat{y}_{m})<y$, Eq. (\ref{eq:12}) can be formulated as follows:
\begin{equation} \label{eq:14}
    \begin{aligned}
    -&\sum^{M}_{m=1}{\|f_{\theta}(\hat{y}_{m})-y\|^{1}_{b_{m}}}=\\
    -&\sum^{M}_{m=1}{\|f_{\theta}(\hat{y}_{m})-(x+n)\|^{1}_{b_{m}}}=\\
    &\sum^{M}_{m=1}{\|x-f_{\theta}(\hat{y}_{m})\|^{1}_{b_{m}}}+\sum^{M}_{m=1}{\|n\|^{1}_{b_{m}}}.
    \end{aligned}
\end{equation}
By considering the last term in Eqs. (\ref{eq:13}) and (\ref{eq:14}), the expectation is:
\begin{equation} \label{eq:15}
    \begin{aligned}
    &\mathbb{E}_{n}\left[\sum^{M}_{m=1}{\|n\|^{1}_{b_{m}}}\right]=\\
    &\mathbb{E}_{n}\left[\sum^{M}_{m=1}{\|(1-b_{m})\odot{n}\|^{1}_{1}}\right]=\\
    &\sum^{M}_{m=1}{\|(1-b_{m})\odot{\mu}\|^{1}_{1}}=\\
    &\sum^{M}_{m=1}{\|(1-b_{m})\odot{0}\|^{1}_{1}}=0.
    \end{aligned}
\end{equation}
Combining Eqs. (\ref{eq:13}) and (\ref{eq:15}) are formulated as follows:
\begin{equation} \label{eq:16}
    \begin{aligned}
    &\mathbb{E}_{n}\left[{\sum^{M}_{m=1}}{\|f_{\theta}(\hat{y}_{m})-\Bar{y}_{m}\|^{1}_{b_{m}}}\right]=\\
    &\mathbb{E}_{n}\left[\sum^{M}_{m=1}{\|f_{\theta}(\hat{y}_{m})-x\|^{1}_{b_{m}}}\right]-\mathbb{E}_{n}\left[\sum^{M}_{m=1}{\|n\|^{1}_{b_{m}}}\right]=\\
    &\sum^{M}_{m=1}{\|f_{\theta}(\hat{y}_{m})-x\|^{1}_{b_{m}}}.
    \end{aligned}
\end{equation}
Moreover, integrating Eqs. (\ref{eq:14}) and (\ref{eq:15}) are computed as follows:
\begin{equation} \label{eq:17}
    \begin{aligned}
    &\mathbb{E}_{n}\left[-{\sum^{M}_{m=1}}{\|f_{\theta}(\hat{y}_{m})-\Bar{y}_{m}\|^{1}_{b_{m}}}\right]=\\
    &\mathbb{E}_{n}\left[\sum^{M}_{m=1}{\|x-f_{\theta}(\hat{y}_{m})\|^{1}_{b_{m}}}\right]+\mathbb{E}_{n}\left[\sum^{M}_{m=1}{\|n\|^{1}_{b_{m}}}\right]=\\
    &\sum^{M}_{m=1}{\|x-f_{\theta}(\hat{y}_{m})\|^{1}_{b_{m}}}.
    \end{aligned}
\end{equation}
Thus, coupling Eqs. (\ref{eq:16}) and (\ref{eq:17}) yields the following:
\begin{equation} \label{eq:18}
    \mathbb{E}_{n}\left[{\sum^{M}_{m=1}}{\|f_{\theta}(\hat{y}_{m})-\Bar{y}_{m}\|^{1}_{b_{m}}}\right]=\begin{cases} 
    {\sum^{M}_{m=1}}{\|f_{\theta}(\hat{y}_{m})-x\|^{1}_{b_{m}}}, \\ \text{$if {\;} f_{\theta}(\hat{y}_{m})>y$}\\ \\
    {\sum^{M}_{m=1}}{\|x-f_{\theta}(\hat{y}_{m})\|^{1}_{b_{m}}}. \\ \text{$otherwise$}
    \end{cases}
\end{equation}
\section{Additions to section 4: Experiment}
\subsection{Synthetic Noise Removal}
\begin{figure*}[h]
\begin{subfigure}{.33\textwidth}
  \centering
  \includegraphics[width=.99\linewidth]{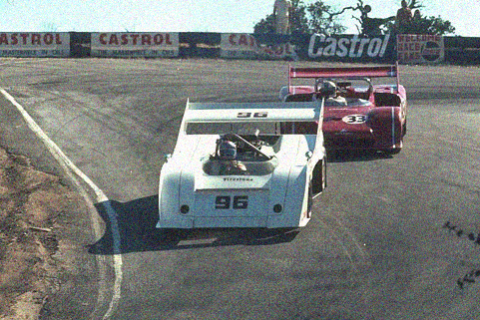}
  \caption{Input}
\end{subfigure}
\begin{subfigure}{.33\textwidth}
  \centering
  \includegraphics[width=.99\linewidth]{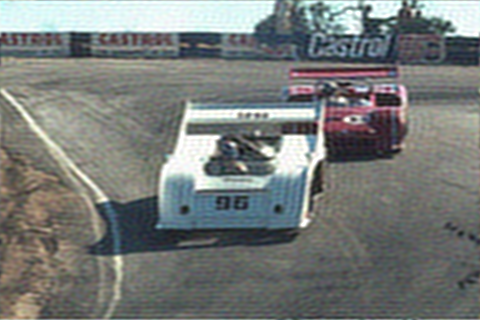}
  \caption{LPF}
\end{subfigure}
\begin{subfigure}{.33\textwidth}
  \centering
  \includegraphics[width=.99\linewidth]{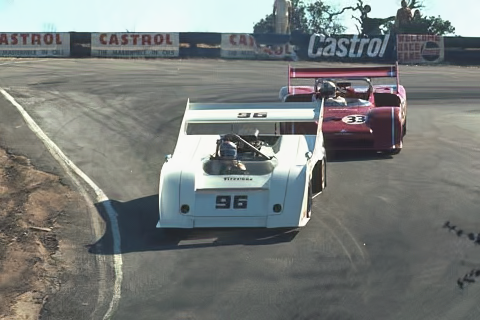}
  \caption{CBM3D}
\end{subfigure}\\
\begin{subfigure}{.33\textwidth}
  \centering
  \includegraphics[width=.99\linewidth]{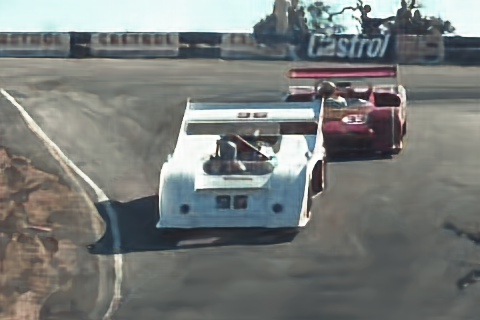}
  \caption{DIP}
\end{subfigure}
\begin{subfigure}{.33\textwidth}
  \centering
  \includegraphics[width=.99\linewidth]{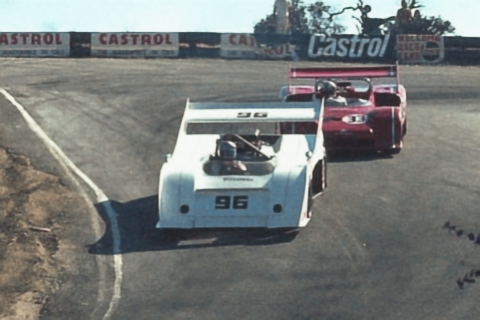}
  \caption{S2S}
\end{subfigure}
\begin{subfigure}{.33\textwidth}
  \centering
  \includegraphics[width=.99\linewidth]{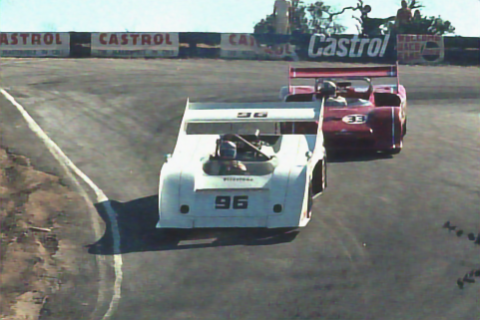}
  \caption{N2F}
\end{subfigure}\\
\begin{subfigure}{.33\textwidth}
  \centering
  \includegraphics[width=.99\linewidth]{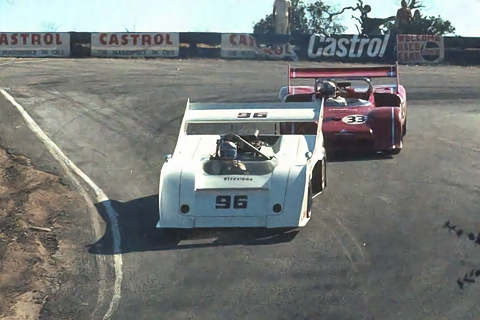}
  \caption{DNCNN}
\end{subfigure}
\begin{subfigure}{.33\textwidth}
  \centering
  \includegraphics[width=.99\linewidth]{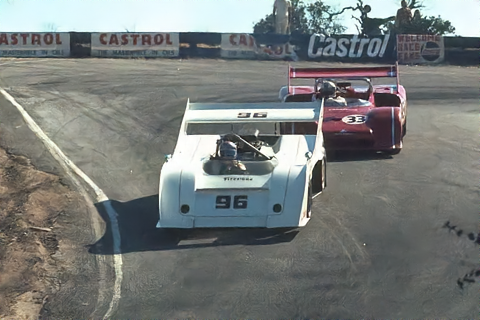}
  \caption{RED30}
\end{subfigure}
\begin{subfigure}{.33\textwidth}
  \centering
  \includegraphics[width=.99\linewidth]{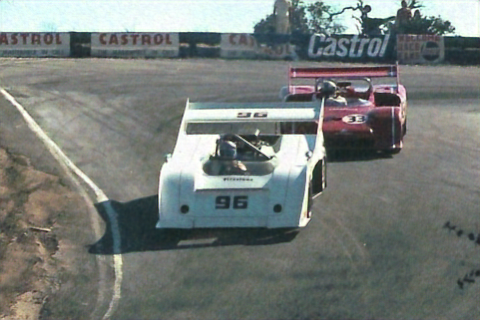}
  \caption{N2V}
\end{subfigure}\\
\begin{subfigure}{.33\textwidth}
  \centering
  \includegraphics[width=.99\linewidth]{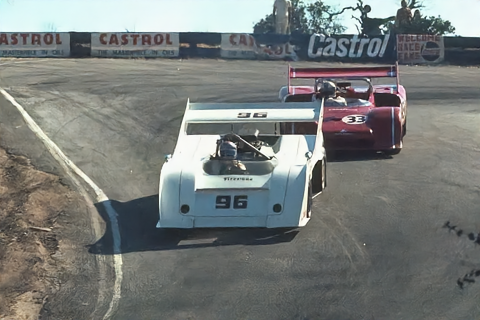}
  \caption{N2N}
\end{subfigure}
\begin{subfigure}{.33\textwidth}
  \centering
  \includegraphics[width=.99\linewidth]{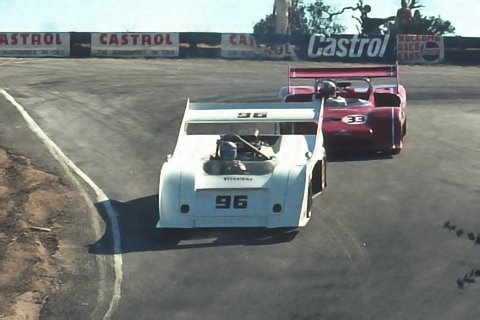}
  \caption{Ours}
\end{subfigure}
\begin{subfigure}{.33\textwidth}
  \centering
  \includegraphics[width=.99\linewidth]{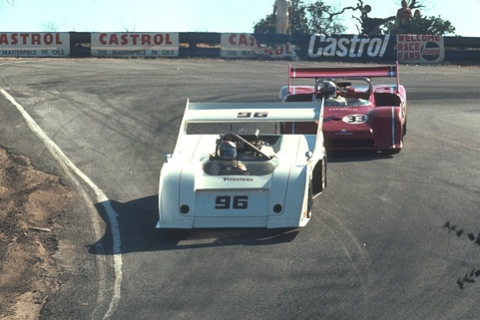}
  \caption{GT}
\end{subfigure}
\caption{Qualitative results of our method and other baselines on \emph{CBSD68} corrupted by AWGN with a noise level $\sigma=15$.}
\label{fig:figure_1}
\end{figure*}
\begin{figure*}[h]
\begin{subfigure}{.33\textwidth}
  \centering
  \includegraphics[width=.99\linewidth]{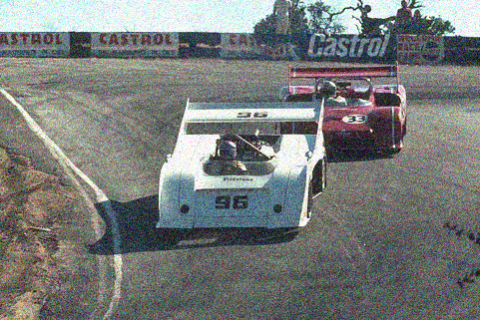}
  \caption{Input}
\end{subfigure}
\begin{subfigure}{.33\textwidth}
  \centering
  \includegraphics[width=.99\linewidth]{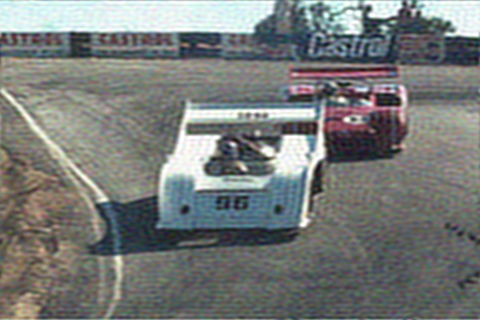}
  \caption{LPF}
\end{subfigure}
\begin{subfigure}{.33\textwidth}
  \centering
  \includegraphics[width=.99\linewidth]{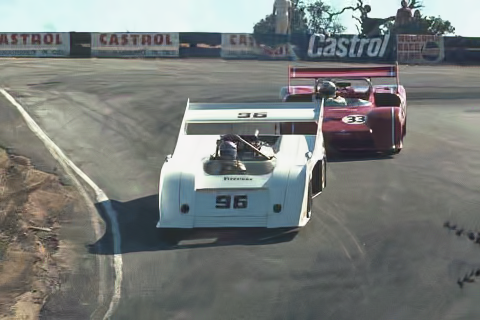}
  \caption{CBM3D}
\end{subfigure}\\
\begin{subfigure}{.33\textwidth}
  \centering
  \includegraphics[width=.99\linewidth]{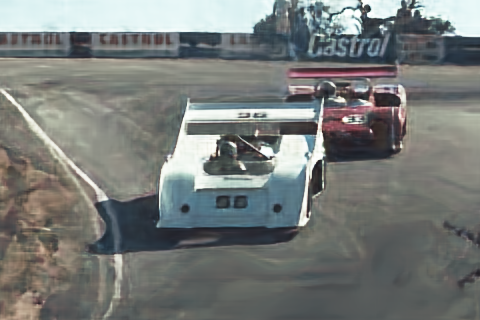}
  \caption{DIP}
\end{subfigure}
\begin{subfigure}{.33\textwidth}
  \centering
  \includegraphics[width=.99\linewidth]{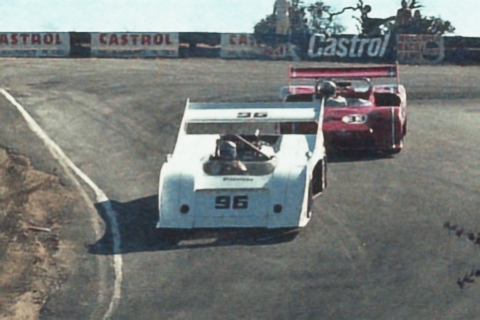}
  \caption{S2S}
\end{subfigure}
\begin{subfigure}{.33\textwidth}
  \centering
  \includegraphics[width=.99\linewidth]{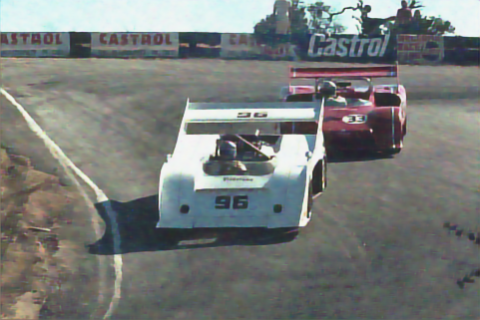}
  \caption{N2F}
\end{subfigure}\\
\begin{subfigure}{.33\textwidth}
  \centering
  \includegraphics[width=.99\linewidth]{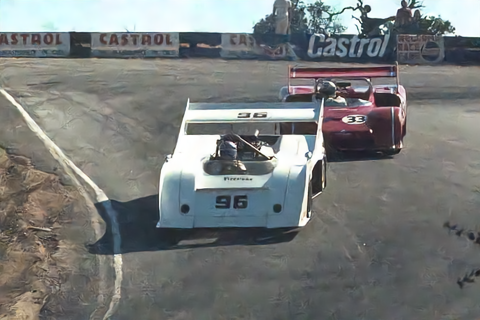}
  \caption{DNCNN}
\end{subfigure}
\begin{subfigure}{.33\textwidth}
  \centering
  \includegraphics[width=.99\linewidth]{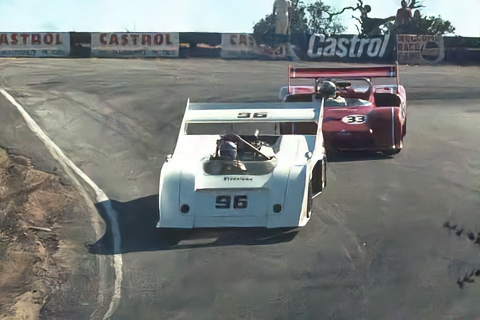}
  \caption{RED30}
\end{subfigure}
\begin{subfigure}{.33\textwidth}
  \centering
  \includegraphics[width=.99\linewidth]{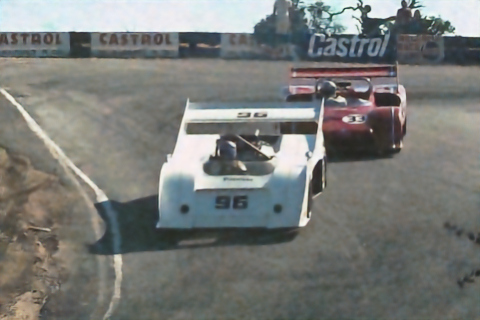}
  \caption{N2V}
\end{subfigure}\\
\begin{subfigure}{.33\textwidth}
  \centering
  \includegraphics[width=.99\linewidth]{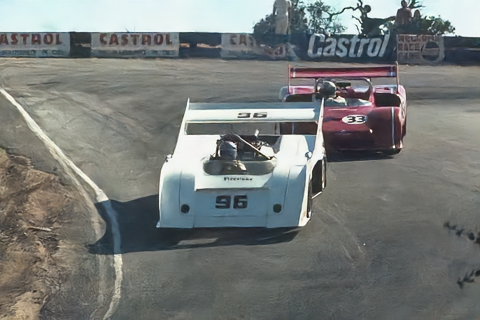}
  \caption{N2N}
\end{subfigure}
\begin{subfigure}{.33\textwidth}
  \centering
  \includegraphics[width=.99\linewidth]{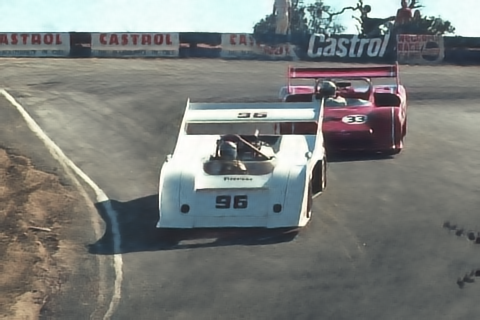}
  \caption{Ours}
\end{subfigure}
\begin{subfigure}{.33\textwidth}
  \centering
  \includegraphics[width=.99\linewidth]{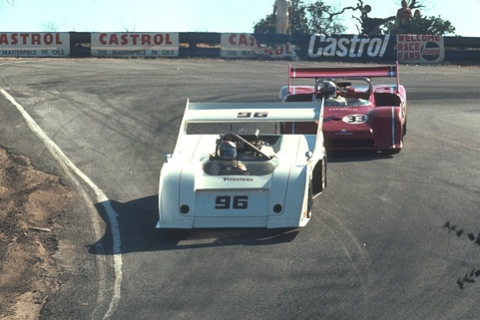}
  \caption{GT}
\end{subfigure}
\caption{Qualitative results of our method and other baselines on \emph{CBSD68} corrupted by AWGN with a noise level $\sigma=25$.}
\label{fig:figure_2}
\end{figure*}
\begin{figure*}[h]
\begin{subfigure}{.33\textwidth}
  \centering
  \includegraphics[width=.99\linewidth]{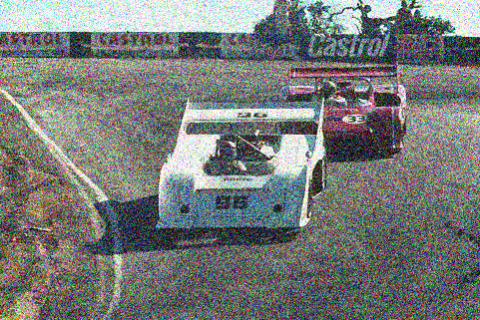}
  \caption{Input}
\end{subfigure}
\begin{subfigure}{.33\textwidth}
  \centering
  \includegraphics[width=.99\linewidth]{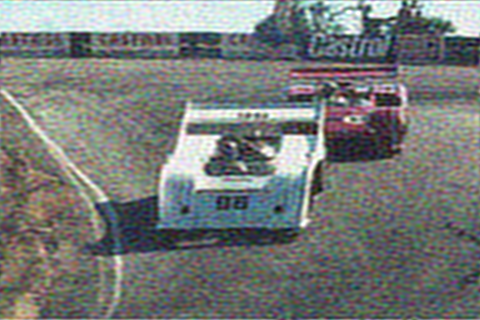}
  \caption{LPF}
\end{subfigure}
\begin{subfigure}{.33\textwidth}
  \centering
  \includegraphics[width=.99\linewidth]{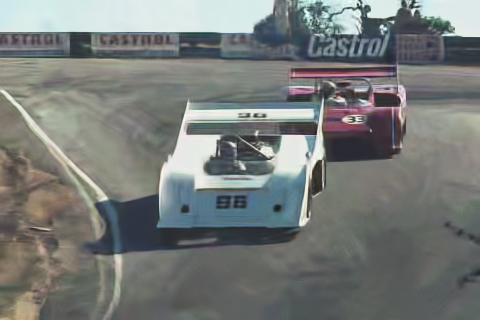}
  \caption{CBM3D}
\end{subfigure}\\
\begin{subfigure}{.33\textwidth}
  \centering
  \includegraphics[width=.99\linewidth]{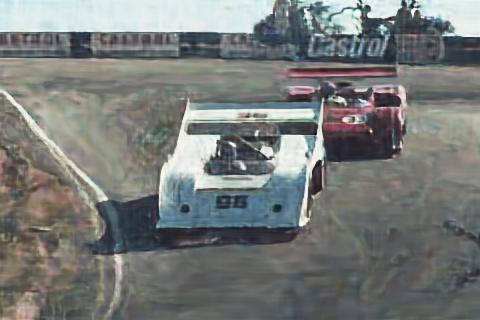}
  \caption{DIP}
\end{subfigure}
\begin{subfigure}{.33\textwidth}
  \centering
  \includegraphics[width=.99\linewidth]{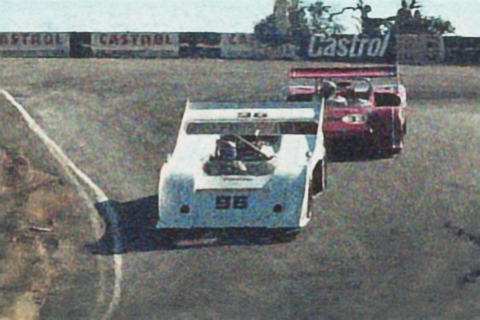}
  \caption{S2S}
\end{subfigure}
\begin{subfigure}{.33\textwidth}
  \centering
  \includegraphics[width=.99\linewidth]{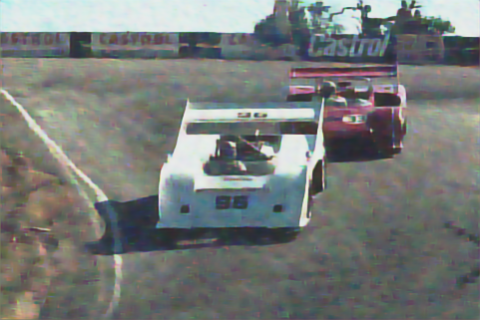}
  \caption{N2F}
\end{subfigure}\\
\begin{subfigure}{.33\textwidth}
  \centering
  \includegraphics[width=.99\linewidth]{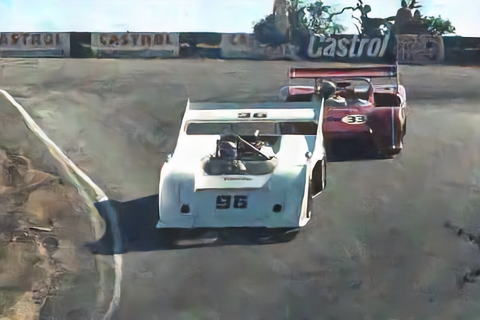}
  \caption{DNCNN}
\end{subfigure}
\begin{subfigure}{.33\textwidth}
  \centering
  \includegraphics[width=.99\linewidth]{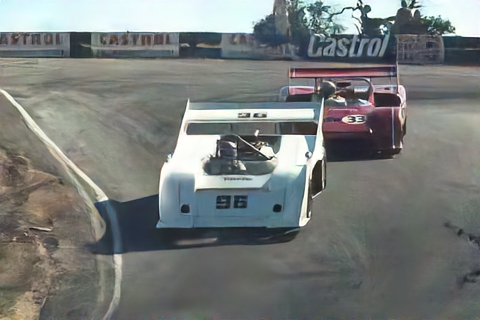}
  \caption{RED30}
\end{subfigure}
\begin{subfigure}{.33\textwidth}
  \centering
  \includegraphics[width=.99\linewidth]{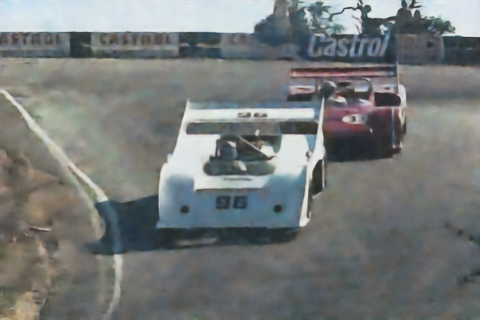}
  \caption{N2V}
\end{subfigure}\\
\begin{subfigure}{.33\textwidth}
  \centering
  \includegraphics[width=.99\linewidth]{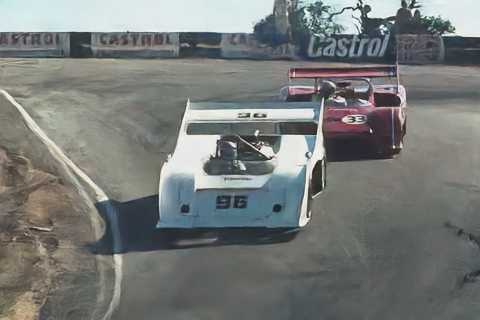}
  \caption{N2N}
\end{subfigure}
\begin{subfigure}{.33\textwidth}
  \centering
  \includegraphics[width=.99\linewidth]{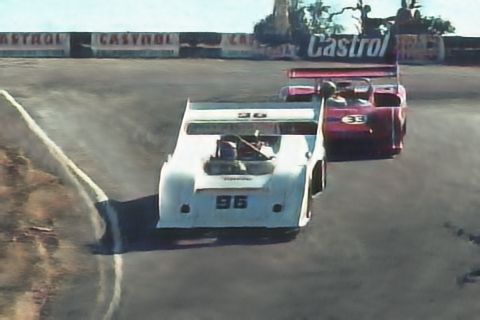}
  \caption{Ours}
\end{subfigure}
\begin{subfigure}{.33\textwidth}
  \centering
  \includegraphics[width=.99\linewidth]{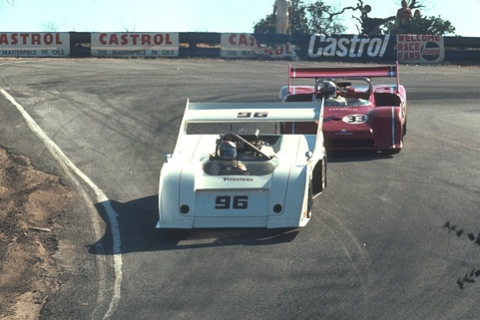}
  \caption{GT}
\end{subfigure}
\caption{Qualitative results of our method and other baselines on \emph{CBSD68} corrupted by AWGN with a noise level $\sigma=50$.}
\label{fig:figure_3}
\end{figure*}
\begin{figure*}[h]
\begin{subfigure}{.245\textwidth}
  \centering
  \includegraphics[width=.99\linewidth]{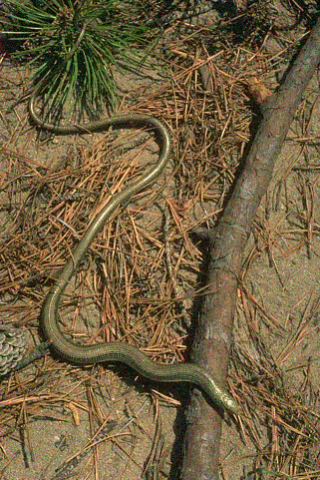}
  \caption{Input}
\end{subfigure}
\begin{subfigure}{.245\textwidth}
  \centering
  \includegraphics[width=.99\linewidth]{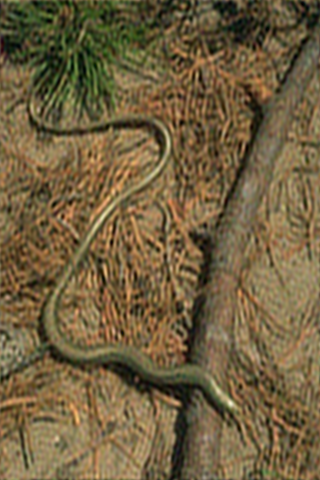}
  \caption{LPF}
\end{subfigure}
\begin{subfigure}{.245\textwidth}
  \centering
  \includegraphics[width=.99\linewidth]{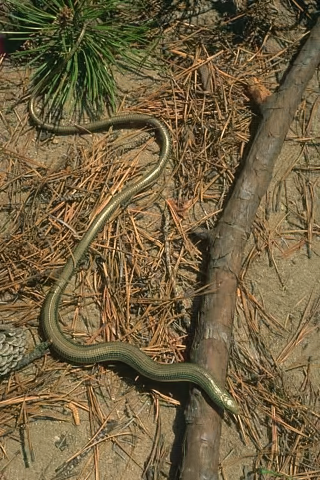}
  \caption{CBM3D}
\end{subfigure}
\begin{subfigure}{.245\textwidth}
  \centering
  \includegraphics[width=.99\linewidth]{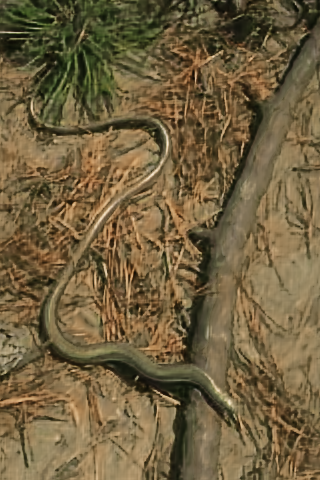}
  \caption{DIP}
\end{subfigure}\\
\begin{subfigure}{.245\textwidth}
  \centering
  \includegraphics[width=.99\linewidth]{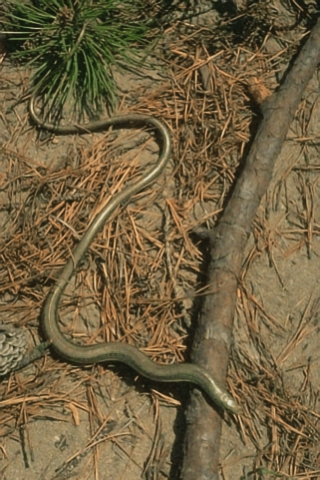}
  \caption{S2S}
\end{subfigure}
\begin{subfigure}{.245\textwidth}
  \centering
  \includegraphics[width=.99\linewidth]{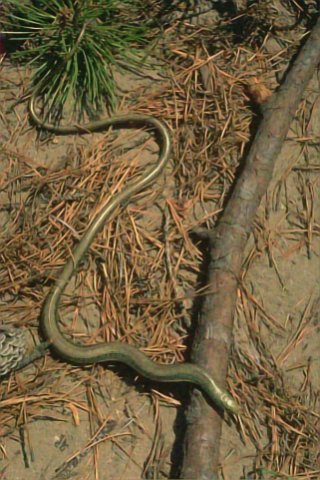}
  \caption{N2F}
\end{subfigure}
\begin{subfigure}{.245\textwidth}
  \centering
  \includegraphics[width=.99\linewidth]{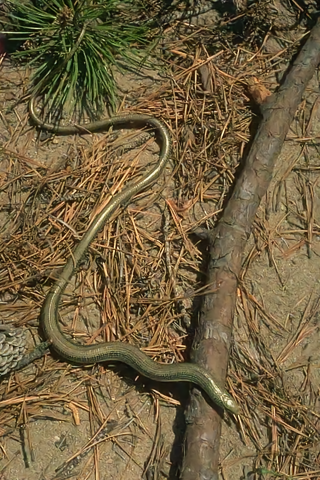}
  \caption{DNCNN}
\end{subfigure}
\begin{subfigure}{.245\textwidth}
  \centering
  \includegraphics[width=.99\linewidth]{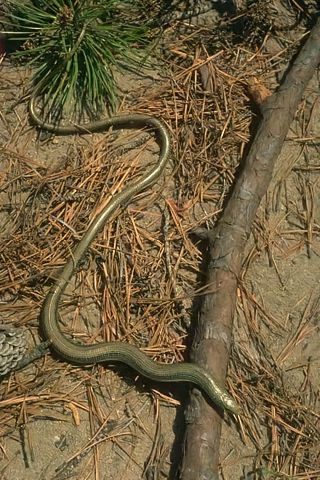}
  \caption{RED30}
\end{subfigure}\\
\begin{subfigure}{.245\textwidth}
  \centering
  \includegraphics[width=.99\linewidth]{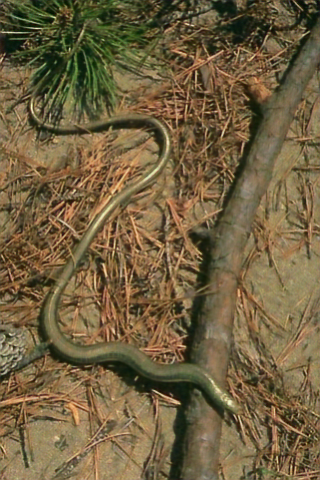}
  \caption{N2V}
\end{subfigure}
\begin{subfigure}{.245\textwidth}
  \centering
  \includegraphics[width=.99\linewidth]{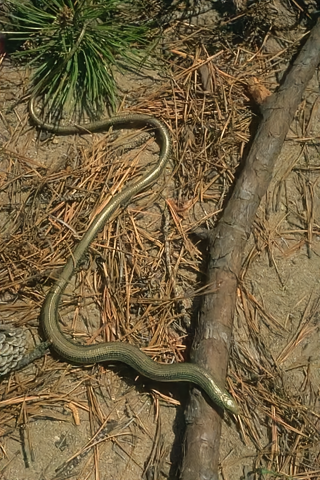}
  \caption{N2N}
\end{subfigure}
\begin{subfigure}{.245\textwidth}
  \centering
  \includegraphics[width=.99\linewidth]{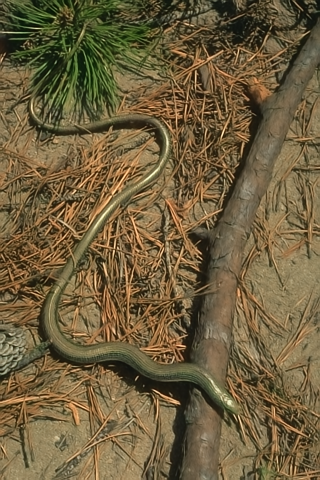}
  \caption{Ours}
\end{subfigure}
\begin{subfigure}{.245\textwidth}
  \centering
  \includegraphics[width=.99\linewidth]{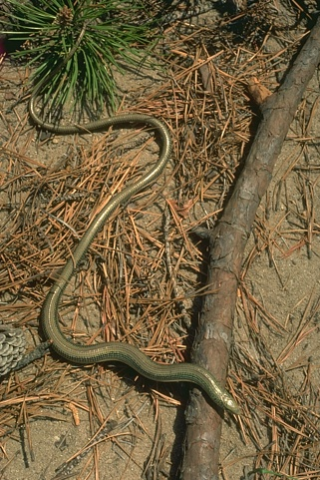}
  \caption{GT}
\end{subfigure}
\caption{Qualitative results of our method and other baselines on \emph{CBSD68} corrupted by AWGN with a noise level $\sigma=15$.}
\label{fig:figure_4}
\end{figure*}
\begin{figure*}[h]
\begin{subfigure}{.245\textwidth}
  \centering
  \includegraphics[width=.99\linewidth]{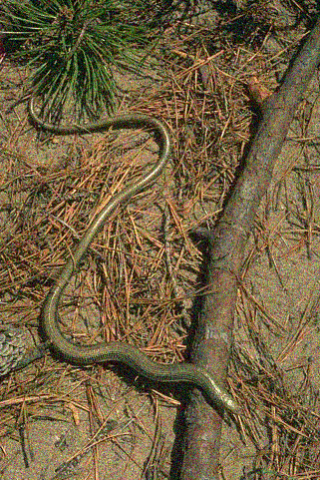}
  \caption{Input}
\end{subfigure}
\begin{subfigure}{.245\textwidth}
  \centering
  \includegraphics[width=.99\linewidth]{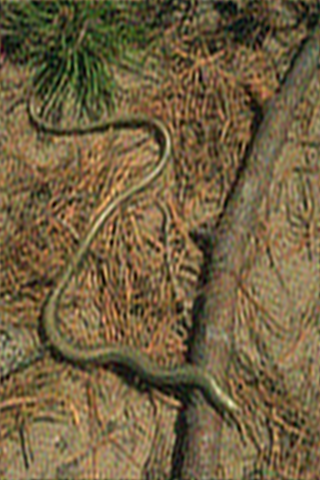}
  \caption{LPF}
\end{subfigure}
\begin{subfigure}{.245\textwidth}
  \centering
  \includegraphics[width=.99\linewidth]{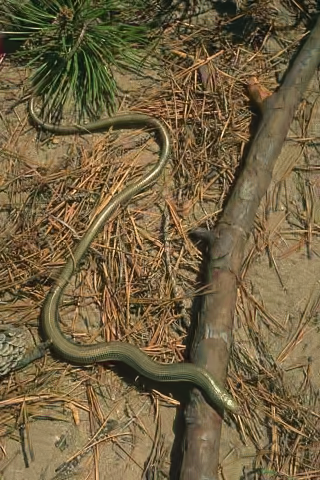}
  \caption{CBM3D}
\end{subfigure}
\begin{subfigure}{.245\textwidth}
  \centering
  \includegraphics[width=.99\linewidth]{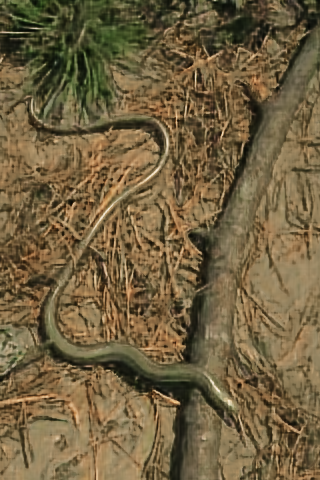}
  \caption{DIP}
\end{subfigure}\\
\begin{subfigure}{.245\textwidth}
  \centering
  \includegraphics[width=.99\linewidth]{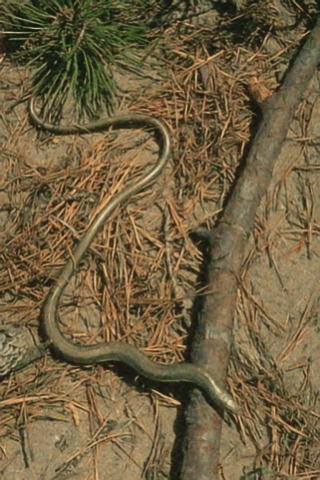}
  \caption{S2S}
\end{subfigure}
\begin{subfigure}{.245\textwidth}
  \centering
  \includegraphics[width=.99\linewidth]{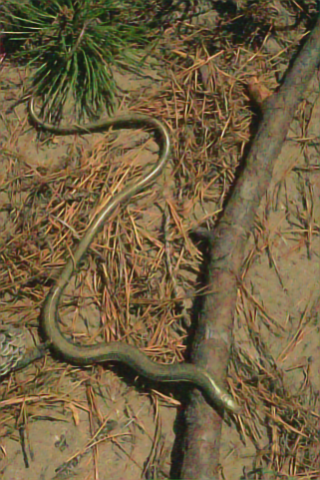}
  \caption{N2F}
\end{subfigure}
\begin{subfigure}{.245\textwidth}
  \centering
  \includegraphics[width=.99\linewidth]{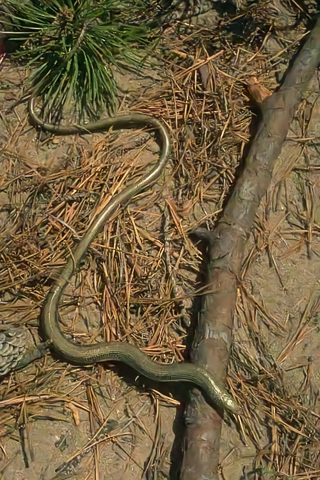}
  \caption{DNCNN}
\end{subfigure}
\begin{subfigure}{.245\textwidth}
  \centering
  \includegraphics[width=.99\linewidth]{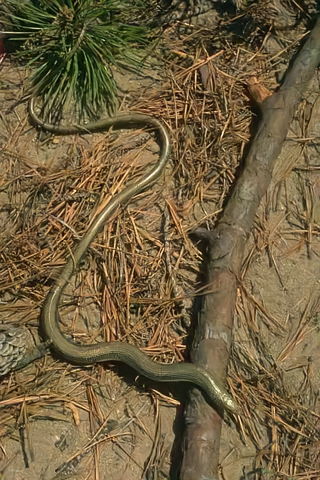}
  \caption{RED30}
\end{subfigure}\\
\begin{subfigure}{.245\textwidth}
  \centering
  \includegraphics[width=.99\linewidth]{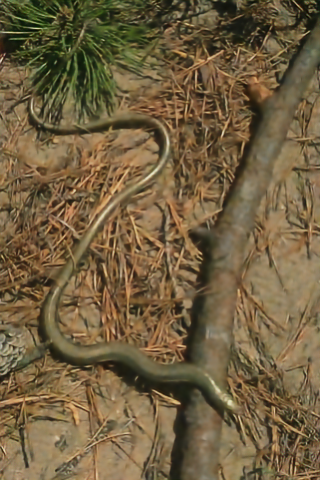}
  \caption{N2V}
\end{subfigure}
\begin{subfigure}{.245\textwidth}
  \centering
  \includegraphics[width=.99\linewidth]{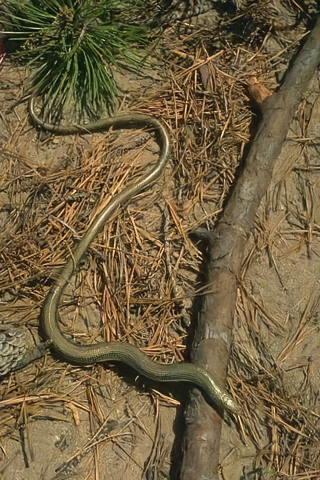}
  \caption{N2N}
\end{subfigure}
\begin{subfigure}{.245\textwidth}
  \centering
  \includegraphics[width=.99\linewidth]{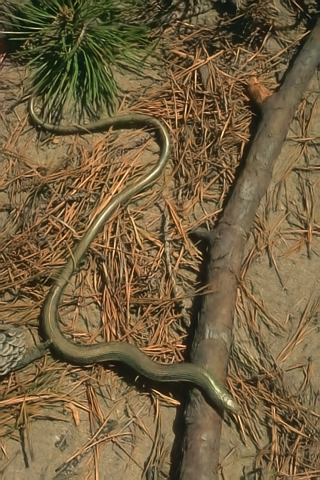}
  \caption{Ours}
\end{subfigure}
\begin{subfigure}{.245\textwidth}
  \centering
  \includegraphics[width=.99\linewidth]{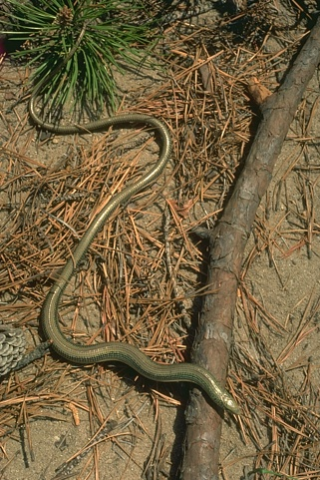}
  \caption{GT}
\end{subfigure}
\caption{Qualitative results of our method and other baselines on \emph{CBSD68} corrupted by AWGN with a noise level $\sigma=25$.}
\label{fig:figure_5}
\end{figure*}
\begin{figure*}[h]
\begin{subfigure}{.245\textwidth}
  \centering
  \includegraphics[width=.99\linewidth]{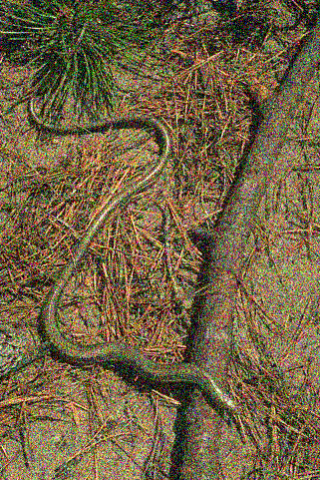}
  \caption{Input}
\end{subfigure}
\begin{subfigure}{.245\textwidth}
  \centering
  \includegraphics[width=.99\linewidth]{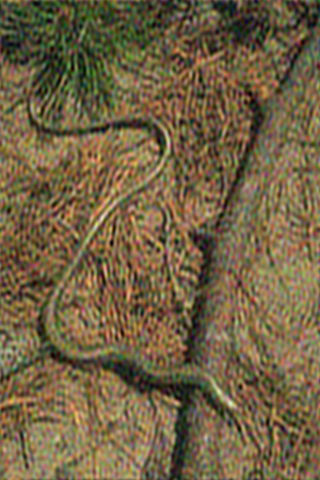}
  \caption{LPF}
\end{subfigure}
\begin{subfigure}{.245\textwidth}
  \centering
  \includegraphics[width=.99\linewidth]{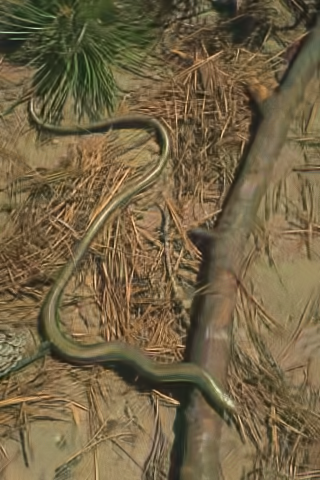}
  \caption{CBM3D}
\end{subfigure}
\begin{subfigure}{.245\textwidth}
  \centering
  \includegraphics[width=.99\linewidth]{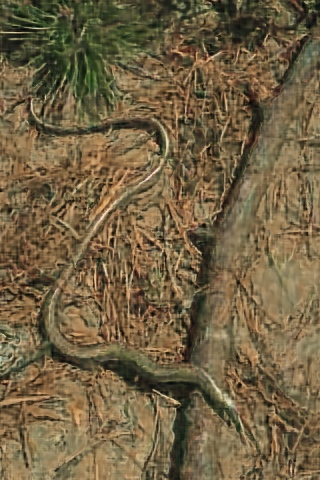}
  \caption{DIP}
\end{subfigure}\\
\begin{subfigure}{.245\textwidth}
  \centering
  \includegraphics[width=.99\linewidth]{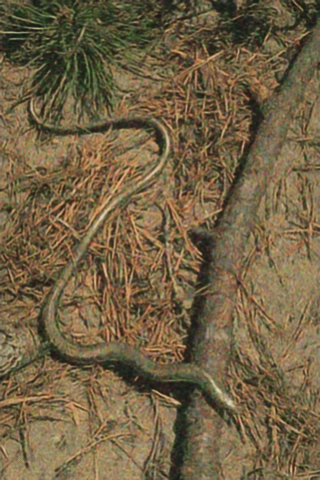}
  \caption{S2S}
\end{subfigure}
\begin{subfigure}{.245\textwidth}
  \centering
  \includegraphics[width=.99\linewidth]{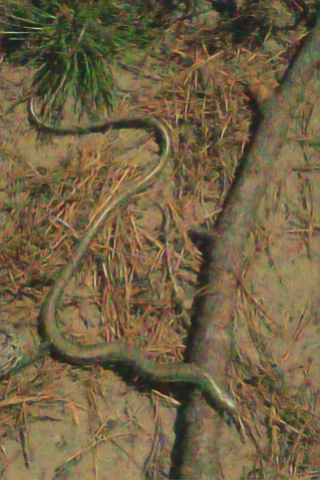}
  \caption{N2F}
\end{subfigure}
\begin{subfigure}{.245\textwidth}
  \centering
  \includegraphics[width=.99\linewidth]{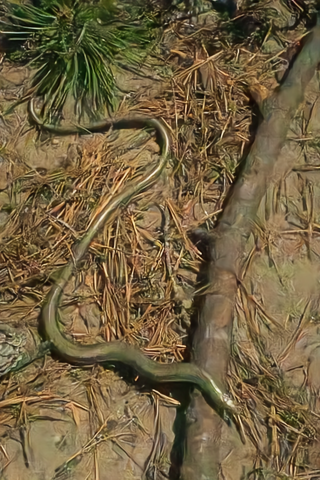}
  \caption{DNCNN}
\end{subfigure}
\begin{subfigure}{.245\textwidth}
  \centering
  \includegraphics[width=.99\linewidth]{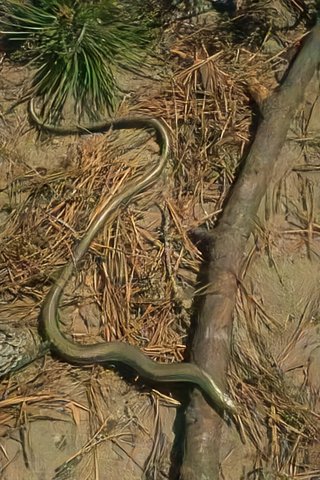}
  \caption{RED30}
\end{subfigure}\\
\begin{subfigure}{.245\textwidth}
  \centering
  \includegraphics[width=.99\linewidth]{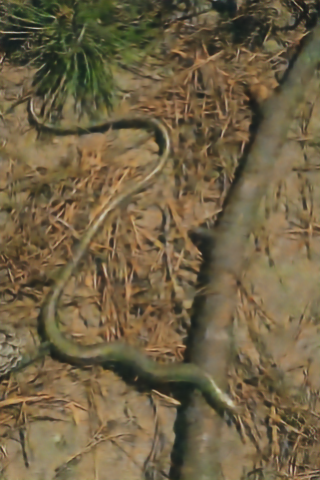}
  \caption{N2V}
\end{subfigure}
\begin{subfigure}{.245\textwidth}
  \centering
  \includegraphics[width=.99\linewidth]{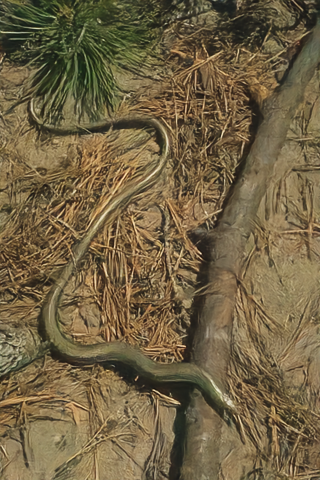}
  \caption{N2N}
\end{subfigure}
\begin{subfigure}{.245\textwidth}
  \centering
  \includegraphics[width=.99\linewidth]{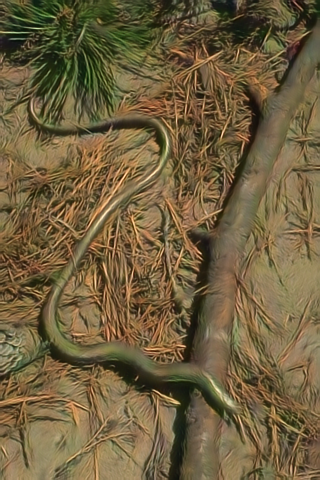}
  \caption{Ours}
\end{subfigure}
\begin{subfigure}{.245\textwidth}
  \centering
  \includegraphics[width=.99\linewidth]{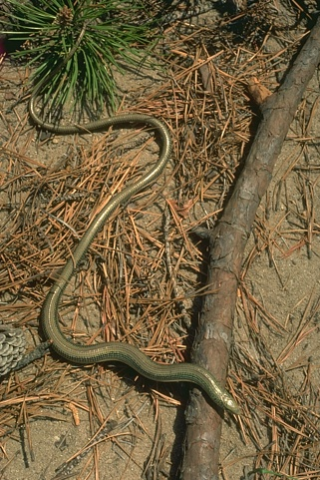}
  \caption{GT}
\end{subfigure}
\caption{Qualitative results of our method and other baselines on \emph{CBSD68} corrupted by AWGN with a noise level $\sigma=50$.}
\label{fig:figure_6}
\end{figure*}
In addition, we present qualitative results on the CBSD68 dataset used in~\cite{cbsd68} with 68 images. Denoising experiments were performed on the additive white Gaussian noise (AWGN) with levels of $\sigma=\{15,25,50\}$. Figure \ref{fig:figure_1}, \ref{fig:figure_2}, \ref{fig:figure_3}, \ref{fig:figure_4}, \ref{fig:figure_5}, and \ref{fig:figure_6} show the qualitative results of our method along with those of selected baselines. Non-learning and single-image learning methods, including LPF, DIP~\cite{dip}, S2S~\cite{s2s}, and N2F~\cite{n2f}, and N2V~\cite{n2v}, a dataset-based deep learning method, left residual noise and excessively smoothed images, with artifacts increasing in frequency along with noise levels. By contrast, our method covered a wide range of noise levels with fewer artifacts and more texture details. Moreover, the proposed method produced images with qualities similar to those of CBM3D~\cite{cbm3d} and N2N~\cite{n2n} with a noise level of $\sigma=50$.
\subsection{Real-World Noise Removal}
\begin{figure*}[h]
\begin{subfigure}{.245\textwidth}
  \centering
  \includegraphics[width=.99\linewidth]{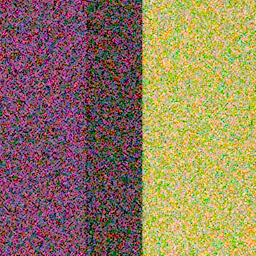}
  \caption{Input}
\end{subfigure}
\begin{subfigure}{.245\textwidth}
  \centering
  \includegraphics[width=.99\linewidth]{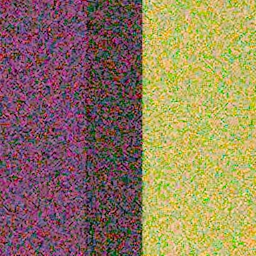}
  \caption{CBM3D}
\end{subfigure}
\begin{subfigure}{.245\textwidth}
  \centering
  \includegraphics[width=.99\linewidth]{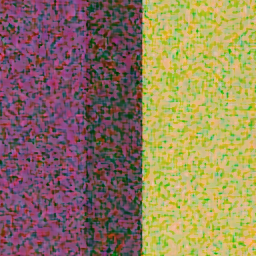}
  \caption{DIP}
\end{subfigure}
\begin{subfigure}{.245\textwidth}
  \centering
  \includegraphics[width=.99\linewidth]{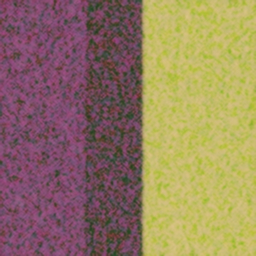}
  \caption{S2S}
\end{subfigure}\\
\begin{subfigure}{.245\textwidth}
  \centering
  \includegraphics[width=.99\linewidth]{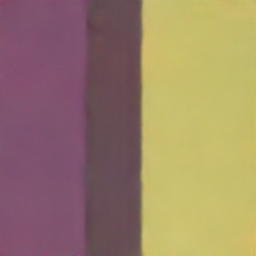}
  \caption{RED30}
\end{subfigure}
\begin{subfigure}{.245\textwidth}
  \centering
  \includegraphics[width=.99\linewidth]{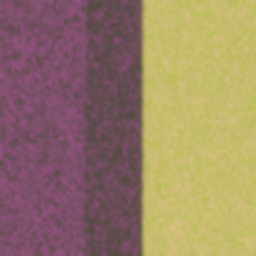}
  \caption{Ours}
\end{subfigure}
\begin{subfigure}{.245\textwidth}
  \centering
  \includegraphics[width=.99\linewidth]{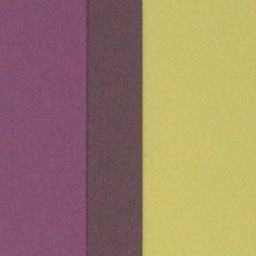}
  \caption{GT}
\end{subfigure}\\
\begin{subfigure}{.245\textwidth}
  \centering
  \includegraphics[width=.99\linewidth]{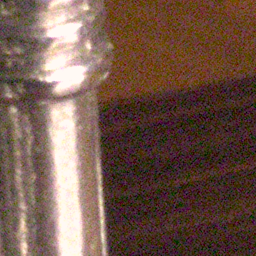}
  \caption{Input}
\end{subfigure}
\begin{subfigure}{.245\textwidth}
  \centering
  \includegraphics[width=.99\linewidth]{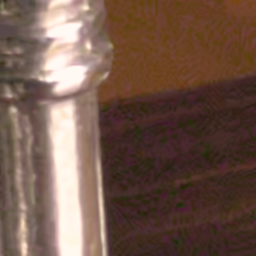}
  \caption{CBM3D}
\end{subfigure}
\begin{subfigure}{.245\textwidth}
  \centering
  \includegraphics[width=.99\linewidth]{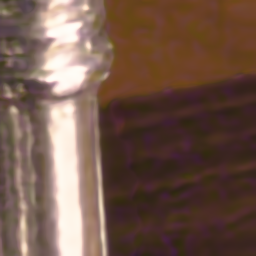}
  \caption{DIP}
\end{subfigure}
\begin{subfigure}{.245\textwidth}
  \centering
  \includegraphics[width=.99\linewidth]{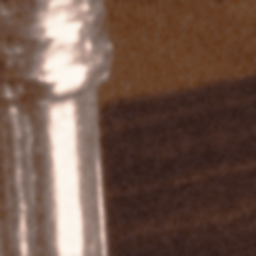}
  \caption{S2S}
\end{subfigure}\\
\begin{subfigure}{.245\textwidth}
  \centering
  \includegraphics[width=.99\linewidth]{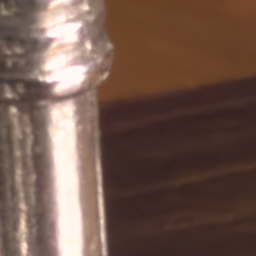}
  \caption{RED30}
\end{subfigure}
\begin{subfigure}{.245\textwidth}
  \centering
  \includegraphics[width=.99\linewidth]{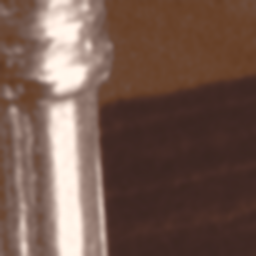}
  \caption{Ours}
\end{subfigure}
\begin{subfigure}{.245\textwidth}
  \centering
  \includegraphics[width=.99\linewidth]{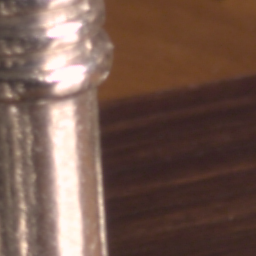}
  \caption{GT}
\end{subfigure}
\caption{Qualitative results of our method and other baselines on \emph{SIDD}.}
\label{fig:figure_7}
\end{figure*}
\begin{figure*}[h]
\begin{subfigure}{.245\textwidth}
  \centering
  \includegraphics[width=.99\linewidth]{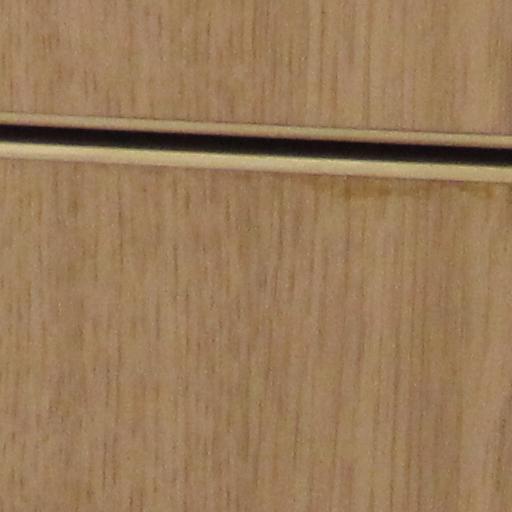}
  \caption{Input}
\end{subfigure}
\begin{subfigure}{.245\textwidth}
  \centering
  \includegraphics[width=.99\linewidth]{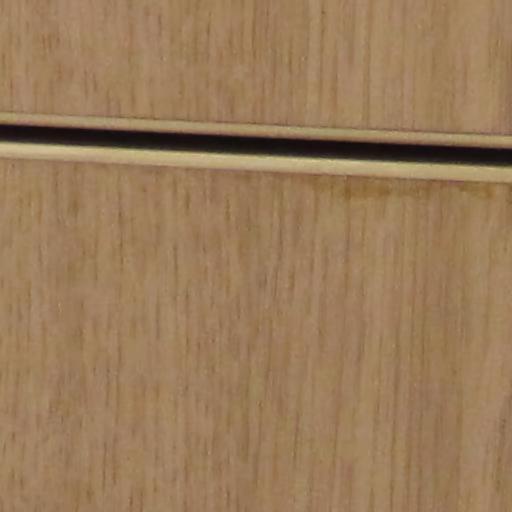}
  \caption{CBM3D}
\end{subfigure}
\begin{subfigure}{.245\textwidth}
  \centering
  \includegraphics[width=.99\linewidth]{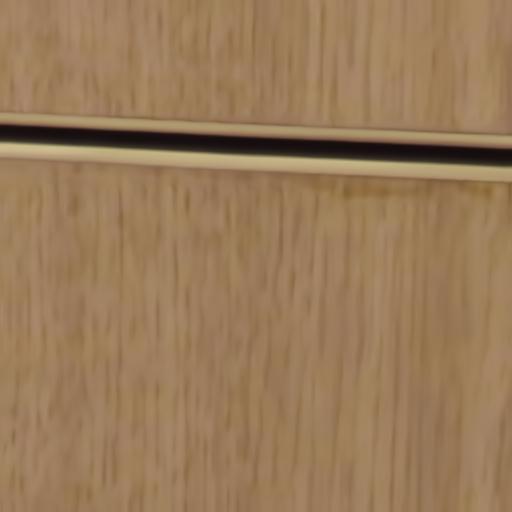}
  \caption{DIP}
\end{subfigure}
\begin{subfigure}{.245\textwidth}
  \centering
  \includegraphics[width=.99\linewidth]{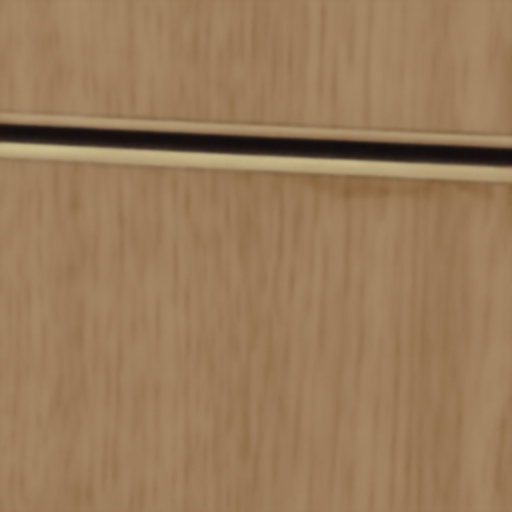}
  \caption{S2S}
\end{subfigure}\\
\begin{subfigure}{.245\textwidth}
  \centering
  \includegraphics[width=.99\linewidth]{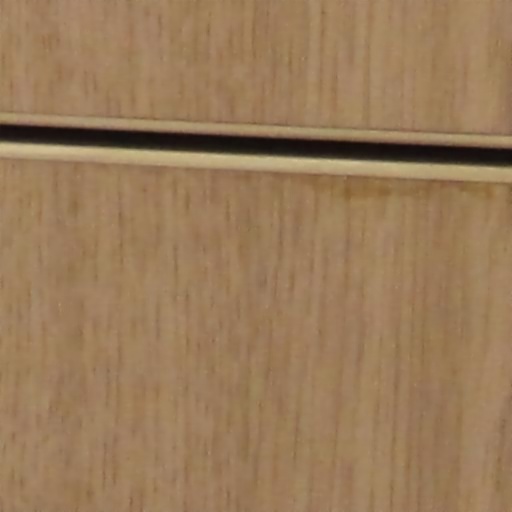}
  \caption{RED30}
\end{subfigure}
\begin{subfigure}{.245\textwidth}
  \centering
  \includegraphics[width=.99\linewidth]{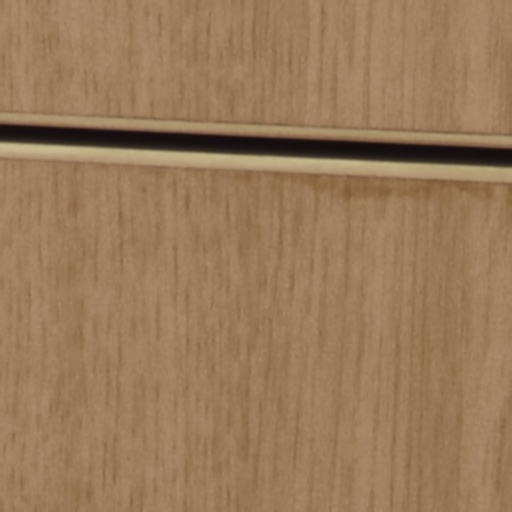}
  \caption{Ours}
\end{subfigure}
\begin{subfigure}{.245\textwidth}
  \centering
  \includegraphics[width=.99\linewidth]{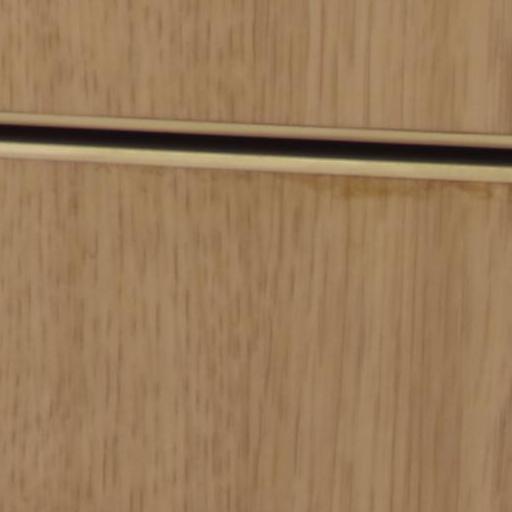}
  \caption{GT}
\end{subfigure}\\
\begin{subfigure}{.245\textwidth}
  \centering
  \includegraphics[width=.99\linewidth]{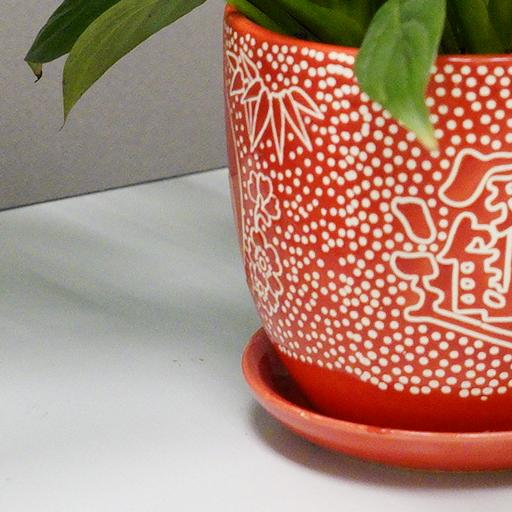}
  \caption{Input}
\end{subfigure}
\begin{subfigure}{.245\textwidth}
  \centering
  \includegraphics[width=.99\linewidth]{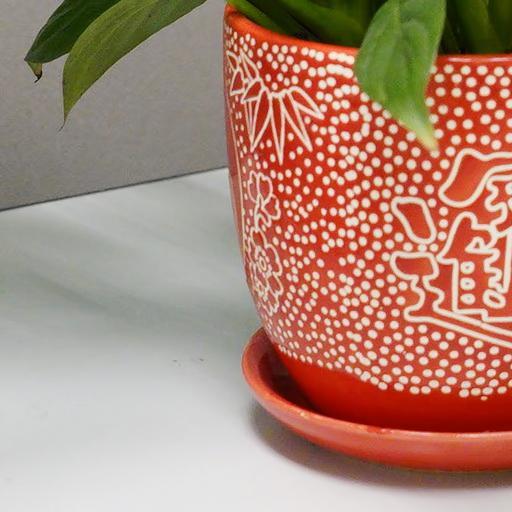}
  \caption{CBM3D}
\end{subfigure}
\begin{subfigure}{.245\textwidth}
  \centering
  \includegraphics[width=.99\linewidth]{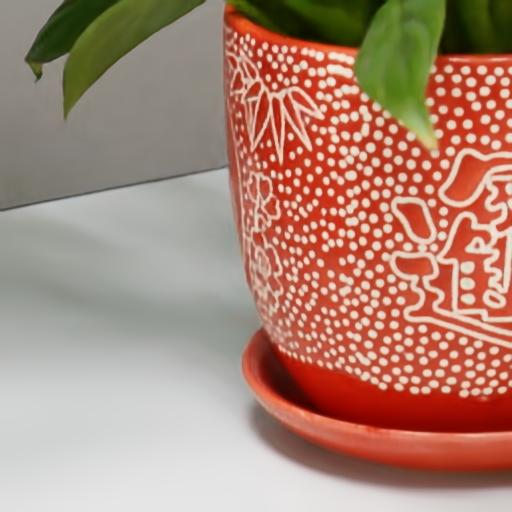}
  \caption{DIP}
\end{subfigure}
\begin{subfigure}{.245\textwidth}
  \centering
  \includegraphics[width=.99\linewidth]{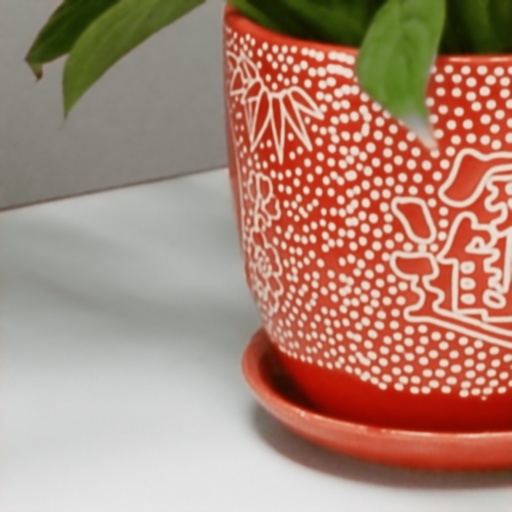}
  \caption{S2S}
\end{subfigure}\\
\begin{subfigure}{.245\textwidth}
  \centering
  \includegraphics[width=.99\linewidth]{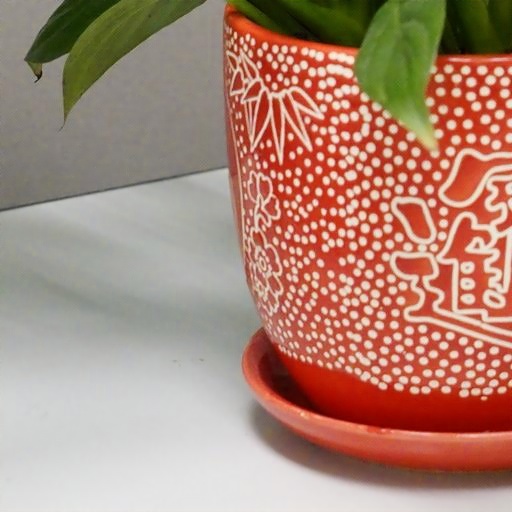}
  \caption{RED30}
\end{subfigure}
\begin{subfigure}{.245\textwidth}
  \centering
  \includegraphics[width=.99\linewidth]{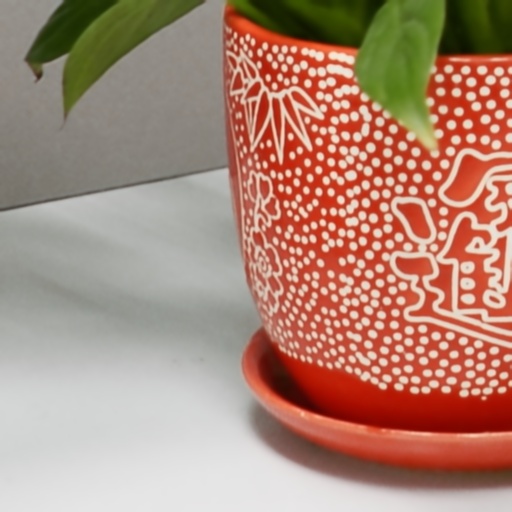}
  \caption{Ours}
\end{subfigure}
\begin{subfigure}{.245\textwidth}
  \centering
  \includegraphics[width=.99\linewidth]{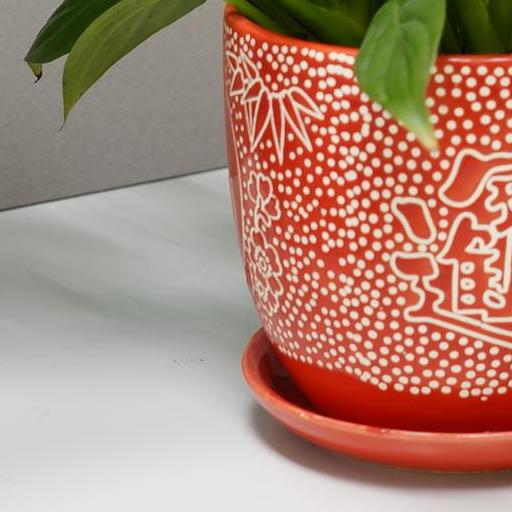}
  \caption{GT}
\end{subfigure}
\caption{Qualitative results of our method and other baselines on \emph{PolyU}.}
\label{fig:figure_8}
\end{figure*}

We also present qualitative results on the real-world noise datasets. 

Qualitative results for the SIDD ~\cite{sidd} are shown in Figure \ref{fig:figure_7}. The CBM3D, DIP, and S2S methods did not sufficiently eliminate noise, exhibiting significant degradation of image quality as well as color inconsistencies. In contrast, our method effectively denoised the images while retaining textural details. 

Qualitative results for the PolyU dataset~\cite{polyu} are shown in Figure \ref{fig:figure_8}. The CBM3D, DIP, and S2S methods did not sufficiently remove noise or significantly smoothed the image. In contrast, our method sufficiently eliminated noise while preserving the lines on the surface.

\bibliography{aaai24}
\end{document}